\documentclass[12pt,a4paper]{article}
\usepackage[utf8]{inputenc}
\usepackage[english]{babel}
\usepackage{indentfirst}
\usepackage{misccorr}
\usepackage{graphicx}
\usepackage{amsmath}
\usepackage{amsthm}
\usepackage{bm}
\usepackage[usenames]{color}
\usepackage{colortbl}

\theoremstyle{definition}

\theoremstyle{plain}

\usepackage[T2A]{fontenc}

\usepackage{float}
\usepackage{caption}
\usepackage{subcaption}

\usepackage{authblk}

\usepackage{amsmath}
\usepackage{amsfonts}
\usepackage{textcomp}
\usepackage{gensymb}

\usepackage{svg}

\usepackage{xcolor}
\usepackage{multirow}
\usepackage{diagbox}
\usepackage{makecell}
\usepackage{etoolbox}
\usepackage[hidelinks]{hyperref}
\usepackage{tabularx}
\newcolumntype{Y}{>{\centering\arraybackslash}X}
\hypersetup{colorlinks=false}

\makeatletter
\renewcommand{\fnum@figure}{Fig. \thefigure}
\makeatother

\title{\bf Unfolder: Fast localization and image rectification of a document with a crease from folding in half}

\author{A.M.~Ershov 1,2, D.V.~Tropin 1,3, E.E.~Limonova 1,3, D.P.~Nikolaev 1,2, V.V.~Arlazarov 1,3\\ 
1 Smart Engines Service LLC, 117312, Moscow, Russia, Prospekt 60-letiia Oktiabria 9\\
2 Institute for Information Transmission Problems of RAS (Kharkevich Institute), 127051, Moscow, Russia, Bolshoy Karetny per. 19, build.1\\ 
3 Federal Research Center "Computer Science and Control" of the Russian Academy of Sciences, 119333, Moscow, Russia, Prospekt 60-letiia Oktiabria 9\\
}

\begin{document}
\maketitle

\begin{center}
	{\bf Abstract}
\end{center}

Presentation of folded documents is not an uncommon case in modern society. Digitizing such documents by capturing them with a smartphone camera can be tricky since a crease can divide the document contents into separate planes. To unfold the document, one could hold the edges potentially obscuring it in a captured image. While there are many geometrical rectification methods, they were usually developed for arbitrary bends and folds. We consider such algorithms and propose a novel approach Unfolder developed specifically for images of documents with a crease from folding in half. Unfolder is robust to projective distortions of the document image and does not fragment the image in the vicinity of a crease after rectification. A new Folded Document Images dataset was created to investigate the rectification accuracy of folded (2, 3, 4, and 8 folds) documents. The dataset includes 1600 images captured when document placed on a table and when held in hand. The Unfolder algorithm allowed for a recognition error rate of 0.33, which is better than the advanced neural network methods DocTr (0.44) and DewarpNet (0.57). The average runtime for Unfolder was only 0.25 s/image on an iPhone XR.


\underline{\it Citation}: Ershov A, Tropin D, Limonova E, Nikolaev D, Arlazarov V. Unfolder: Fast localization and image rectification of a document with a crease from folding in half. Computer Optics 20XX; 4X(X): XXX-YYY. DOI: 10.18287/2412-6179-CO-editorial index.

\begin{center}
	{\bf Introduction}
\end{center}

\begin{figure}[ht!]
\begin{minipage}[b]{0.265\linewidth}
  \centering
  \centerline{\includegraphics[width=\textwidth]{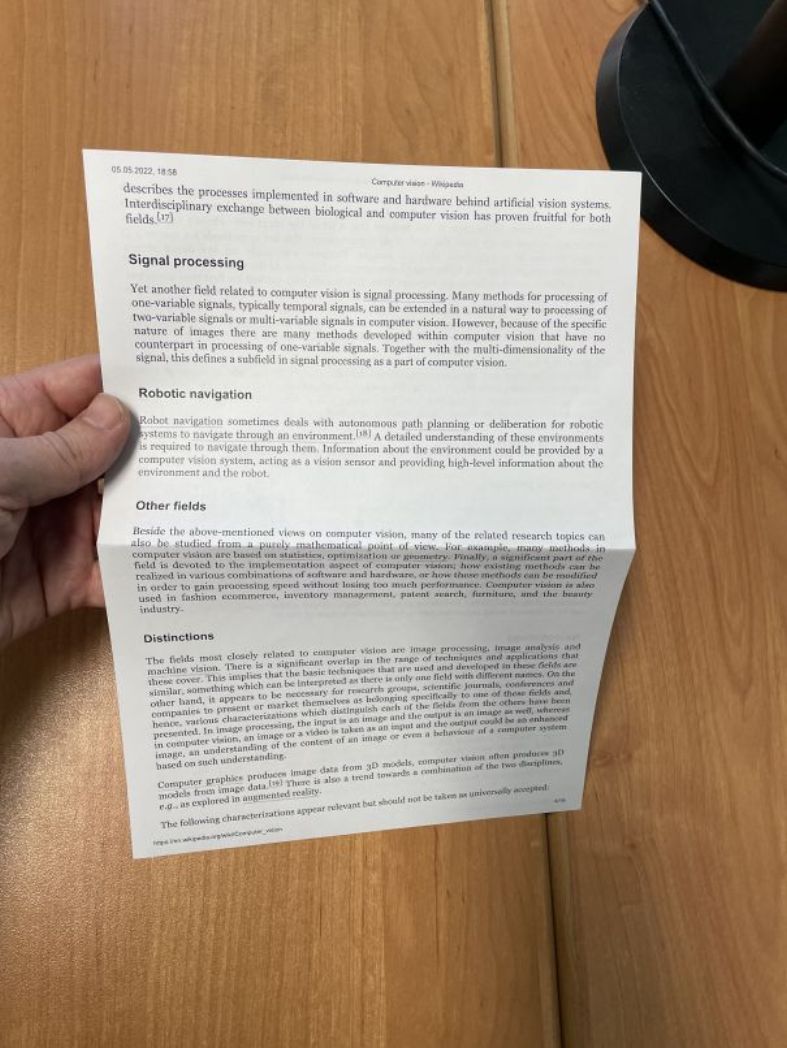}}
  \centerline{(a)}
\end{minipage}
\hfill
\begin{minipage}[b]{0.25\linewidth}
  \centering
  \centerline{\includegraphics[width=\textwidth]{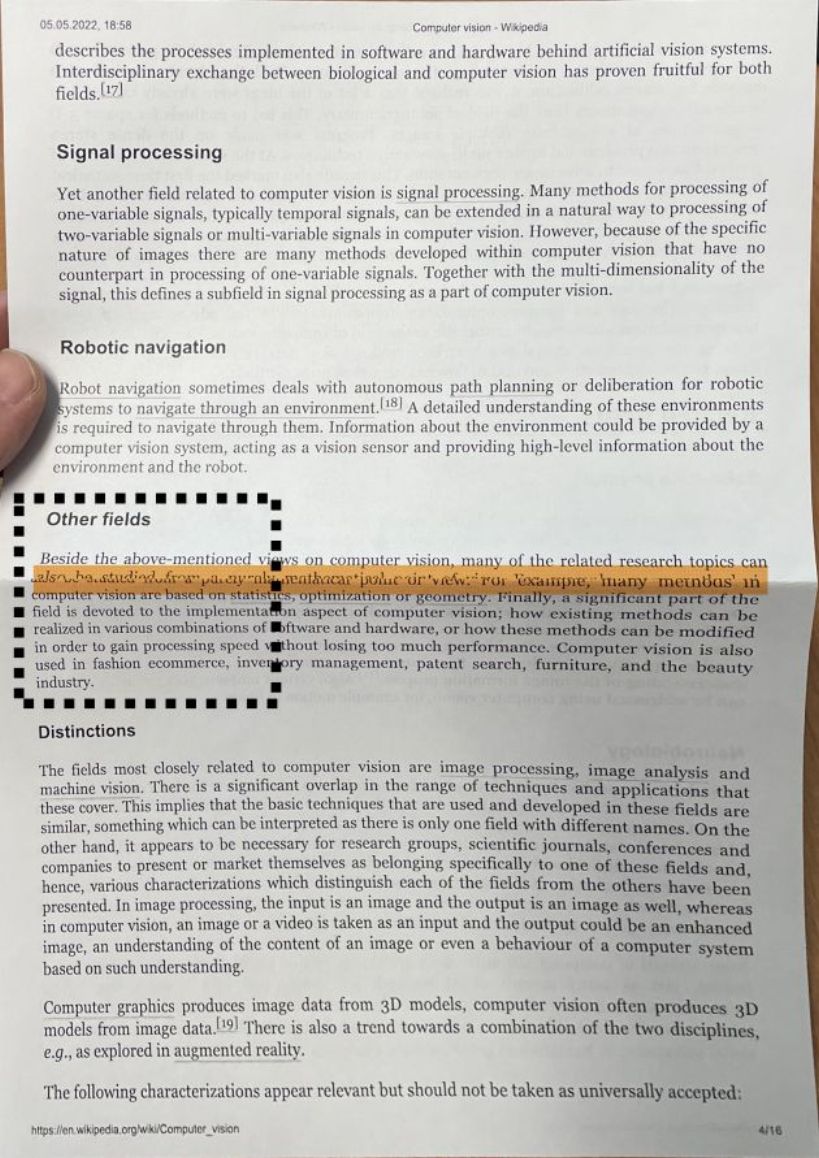}}
  \centerline{(b)}
\end{minipage}
\hfill
\begin{minipage}[b]{0.438\linewidth}
  \centering
  \centerline{\includegraphics[width=\textwidth]{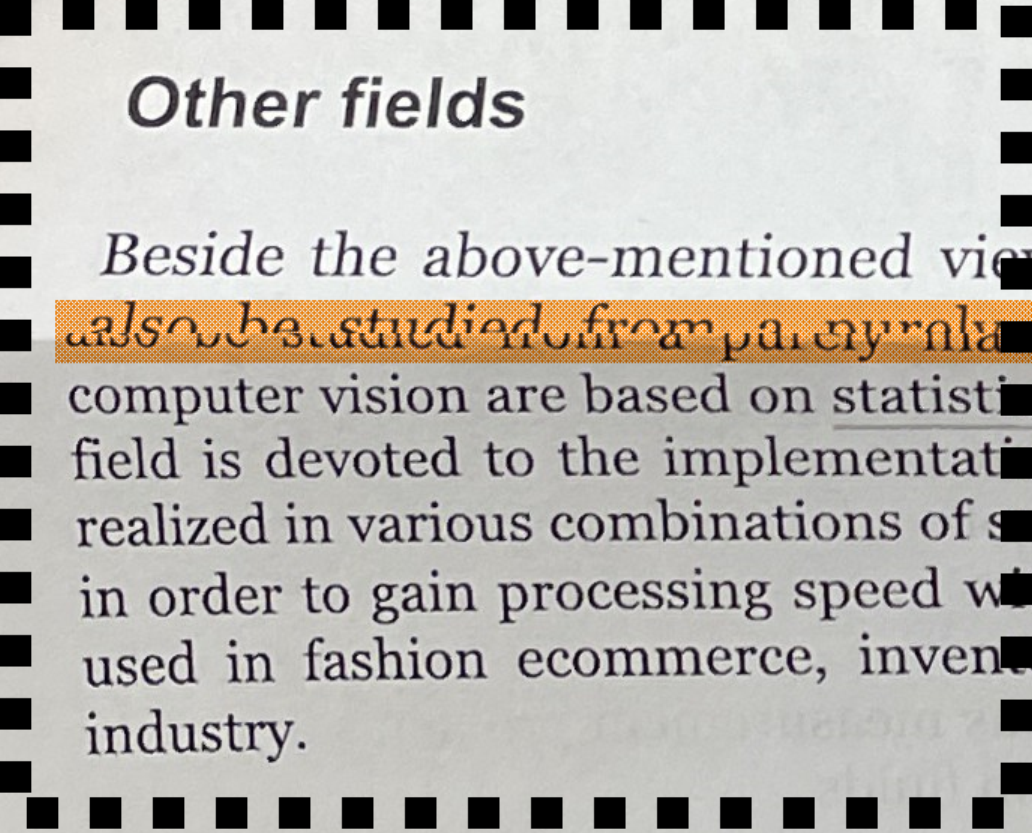}}
  \centerline{(c)}
\end{minipage}
\caption{(a) Input image, (b) rectified image with a content tearing (marked in orange) caused by inappropriate image stitching, (c) enlarged region. }
\label{fig:fragmentation}
\end{figure}

Physical documents digitalization has become a crucial task in the modern world.
This problem used to be solved with the help of flatbed scanners with further Optical Character Recognition (OCR)[1].
Nowadays smartphones have become widely-spread, so the camera-captured images are no less popular than the scanned ones.
However, common OCR methods can not be performed on the camera-captured images straightforwardly as the document can be heavily distorted.
For example, a document can be folded in half and presented to the camera as shown on Fig.~\ref{fig:fragmentation}.\textit{a}.
Thus an additional step --- image rectification [2-5] is required.
The rectification includes restoring a document image as if it was scanned with a flatbed scanner based on the image of a document captured with a mobile device (see Fig.~\ref{fig:fragmentation}.\textit{b}).
The creases could complicate the rectification, and then the problem is referred to as image dewarping, image unwarping, image restoration, textures unwrapping.

We found a single work among the published studies dedicated to rectification which considers the documents folded in half. S. Das et al. in [6] proposed to localize the contour of the document parts (defined by the crease), then rectify each part independently, and concatenate them in a resulting image.
However, the authors themselves stated that the document content could be torn (see Fig.~\ref{fig:fragmentation}.\textit{b} and~\ref{fig:fragmentation}.\textit{c} which illustrate the tearing of the content) in the concatenation areas of the rectified image. 

This work examines the available rectification algorithms and provides details on the developed algorithm. We designed the proposed solution to (i) rectify an image of a document with a crease from folding in half which was captured with a smartphone camera, (ii) guarantee the absence of content tearing, and (iii) be executable on smartphones.

One of the difficulties when developing such an algorithm is the absence of a sufficient dataset.
In [6], only 9 images are published.
The DocUNet~[7] dataset includes about 10 images of documents with a crease from folding in half.
The recently published WarpDoc~[8] dataset includes a subset of folded documents, but most of the creases are arbitrary, often not dividing the document in half and non-parallel to its edges.


In this paper a new dataset FDI is presented.
In contrast with the aforementioned ones it is not designed to fulfill all possible document warps or arbitrary folds, concentrating only on specific, but very common cases.
This dataset contains multiple types of folds (2, 3, 4, 8) of different documents when the document placed on a table and when held in hand.

The main contributions of the present paper are the following:
\begin{enumerate}
\item We propose an algorithm Unfolder, being able to rectify images of documents folded in half. The Unfolder allows to rectify images very fast even on smartphone central processors (0.25 s on iPhone XR).
\item We present a criterion of continuity of a mapping comprised of two projective transformations used in the Unfolder algorithm. It makes possible to get rid of content tearing on the line of image concatenation.
\item We introduce a brand-new dataset FDI comprised of real camera-captured images of folded documents.
\item The Unfolder sets an initial baseline performance on the twofold subset of FDI, outperforming the current state-of-the-art algorithms.
\end{enumerate}

The paper is structured as follows: Section 1 reviews published rectification methods, the absolute majority of which does not consider the specific case of folded documents; Section 2 explains the proposed Unfolder algorithm in detail; Section 3 describes the proposed Folded Document Images (FDI) dataset, provides the measurement of the accuracy and runtime for the Unfolder algorithm, and compares its performance with other methods. 


\begin{center}
	{\bf 1. Related work}
\end{center}
Certain rectification methods explicitly reconstruct the surface of a document in three-dimensional space. 
They employ shape from \textit{x}, where \textit{x} is shading [9, 10], motion [11, 12], and some methods rely on additional equipment such as stereo cameras or lasers [13--16].
Such methods are either computationally expensive or require multiple captured images or additional equipment.

Another approach (e.g., [6, 17--19]) to the rectification problem includes two steps: document contour detection, and document image normalization using the Coons patch [20].
The document image normalization, however, could be inappropriate -- when normalizing projectively distorted documents, the document's contour is corrected, but its content is distorted.

Another way to rectify the image given the borders of the document is to make assumptions on the model of the document in the image.
The document surface can be supposed to be cylindrical [21--23].
Additionally, the document content is used to help the rectification.
The authors make use of the text lines to improve their methods.
These constraints make it possible to simplify the rectification, though are not suitable for all possible images.
There are works both not including projectivity in such a case [21] and including [22, 23].
In all these works a grid is generated on the document to make the rectification possible.
Modern researchers focus on the neural network-based approach to the rectification problem [7, 8, 24--40].
The DocUNet [7] dataset has been adopted for benchmarking.
To compare the performance of the proposed algorithm with the published algorithms, we consider DewarpNet [24]. Several authors of the latter have introduced the DocUNet dataset in the first place.
We also consider the recent state-of-the-art (among the published studies) neural network method, DocTr~[26].
The authors of corresponding papers provided open source code as well as the pre-trained neural network models.

Among the algorithms we have considered, none would allow the uncorrupted rectification of a document folded in a half using only one frame and take less than a second on a  smartphone mobile processor. 
Therefore, in the next section we propose a new algorithm Unfolder.


\begin{center}
	{\bf 2. Proposed algorithm}
\end{center}

\begin{figure}[ht!]
\centering
\includegraphics[width=\linewidth]{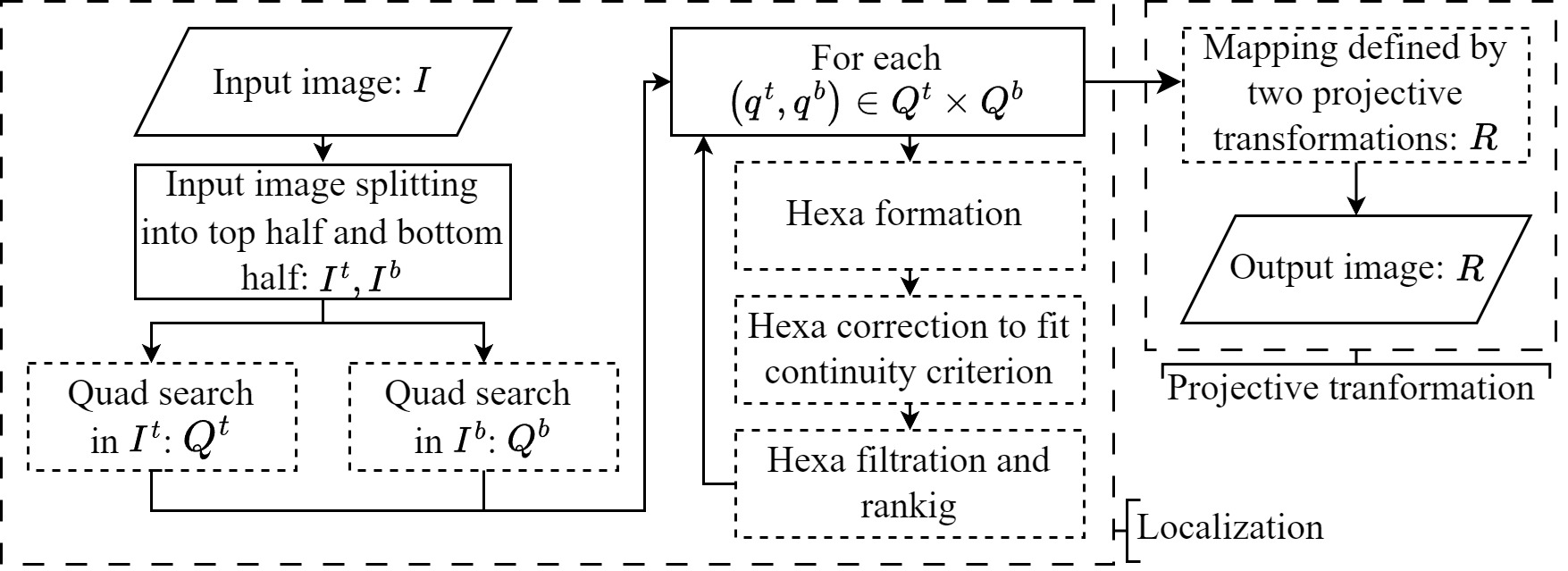}
\caption{Schematic structure of the Unfolder algorithm}
\label{fig:block}
\end{figure}

In this section we propose the new algorithm Unfolder for rectification of twofold document image.
We suppose that the document on the image consists of two concatenated flat rectangles.
So localization of the quadrilaterals of the outer contour and projective transformation for each of them are two main steps of Unfolder (see Fig. \ref{fig:block}).

Consider the algorithm scheme from Fig.~\ref{fig:block} in detail.
We suppose the upper half of the input image to contain most of the upper half of the document, and the bottom half of the image to contain most of the bottom half of the document.
So at first the input image $I$ is split into top half $I^t$ and bottom half $I^b$, and we try to roughly approximate the halves of the documents by two quadrilaterals.
Thereby, both halves are subjected to an algorithm of quadrilateral localization based on edges and lines extraction [5] having a set of quadrilaterals 
as its output.
Thus sets of quadrilaterals $Q^t$ from $I^t$ and $Q^b$ from $I^b$ are extracted (see Sec.~2.1 for more details).

Each pair of quadrilaterals is used to form a hexangle representing the document outer borders on the image $I$. 
For that we develop special refining techniques (see Sec.~2.2).
All the generated hexangles undergo a correction step to fit a continuity criterion (see Sec.~2.3), so after the projective transformation there is no the content tearing.
After that the hexangles are sorted by their \textit{contour score} and the one with the highest \textit{contour score} is chosen as an output of the localization step (see Sec.~2.4). 

The final step of Unfolder is a mapping defined by two projective transformations of the input image (see Sec.~2.5), so the rectified image is an output of the proposed algorithm.
Let us consider all the dashed blocks of the scheme in detail.

\begin{center}
	\underline{2.1 Quadrilateral search}
\end{center}
\label{subsec:quads}
An algorithm proposed in [5] is employed as follows.
At first, the images $E_h, E_v$ of horizontal and vertical edges are extracted the same way as in~[5].
Let us denote $L_h$ all lines, hereinafter referenced as \textit{horizontal}, having an angle with the horizontal directing vector lying in the interval $(-\frac{\pi}{4}, \frac{\pi}{4})$.
Similarly $L_v$ as a set of \textit{vertical} lines, having angle with the vertical direvting vector lying in the interval $(-\frac{\pi}{4}, \frac{\pi}{4})$ is denoted.
The sets $L_h$ and $L_v$ are localized utilizing Fast Hough Transform~[41].
Let us denote the line separating $I$ into halves as $l_s$.
The desired quadrilaterals are formed by intersecting four lines each: one \textit{horizontal} line $l_h \in L_h$, the line $l_s$ and two \textit{vertical} lines $l_v^{(1, 2)} \in L_v$.
All possible quadrilaterals are localized by the brute-force search among all lines configurations and divided into those lying fully in $I^t$ and those lying fully in $I^b$.
As a result we get two sets of all quadrilaterals: $Q^t = \{q^t \subset I^t\}$ and $Q^b = \{q^b \subset I^b\}$.
These sets are separately sorted with the \textit{contour score} described in [5].

\begin{figure}[ht!]
\centering
\begin{minipage}[b]{0.29\linewidth}
  \centering
  \centerline{\includegraphics[width=\textwidth]{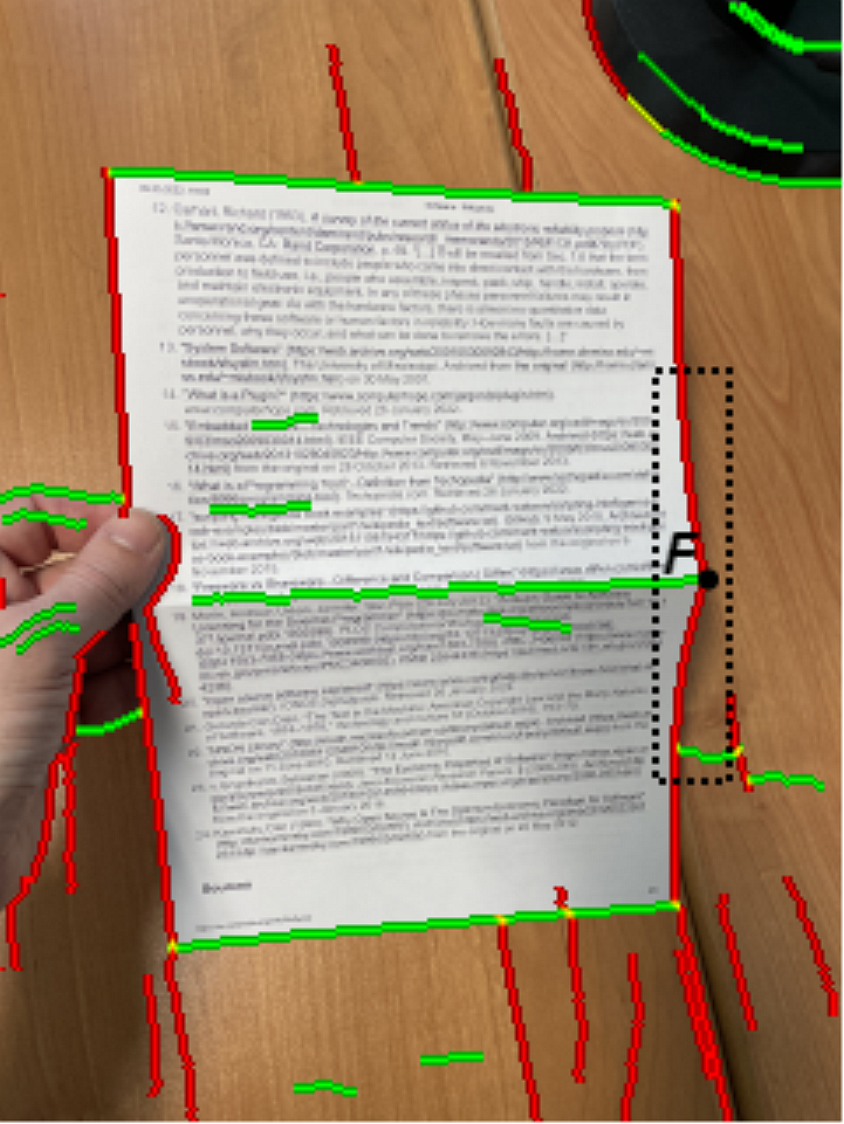}}
  \centerline{(a) }
\end{minipage}
\hfill
\begin{minipage}[b]{0.29\linewidth}
  \centering
  \centerline{\includegraphics[width=\textwidth]{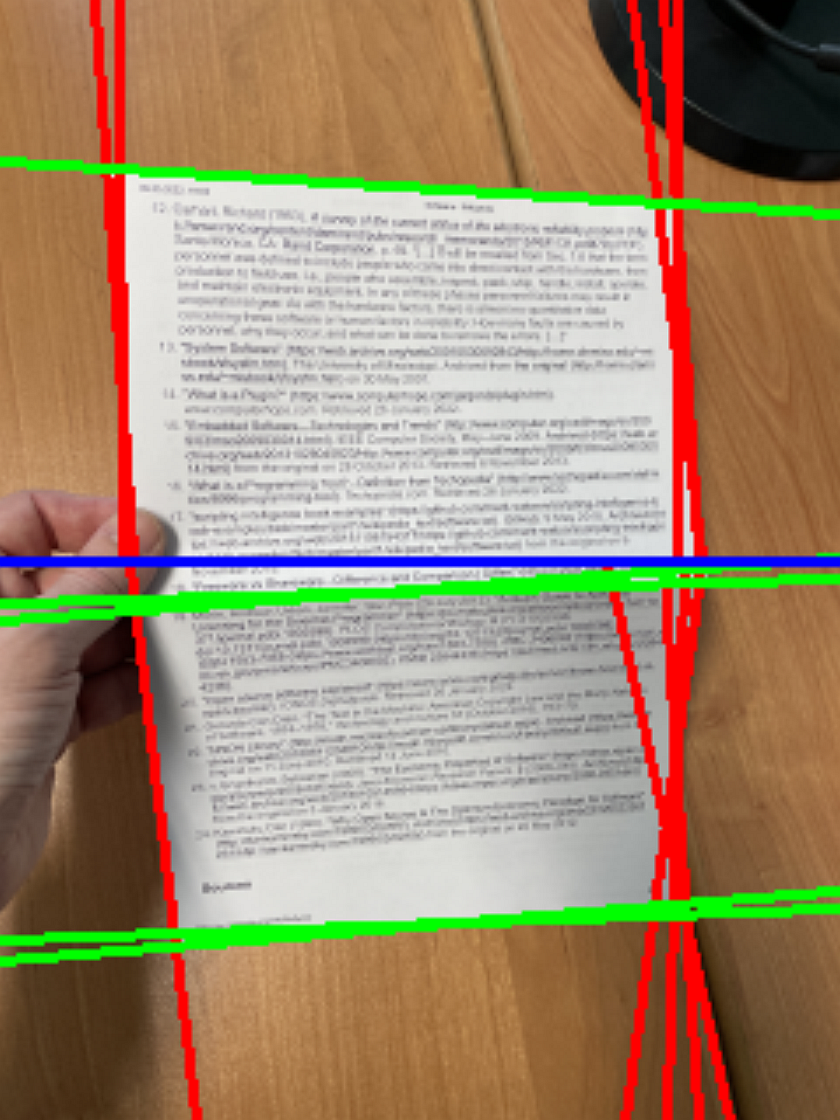}}
  \centerline{(b)}
\end{minipage}
\hfill
\begin{minipage}[b]{0.29\linewidth}
  \centering
  \centerline{\includegraphics[width=\textwidth]{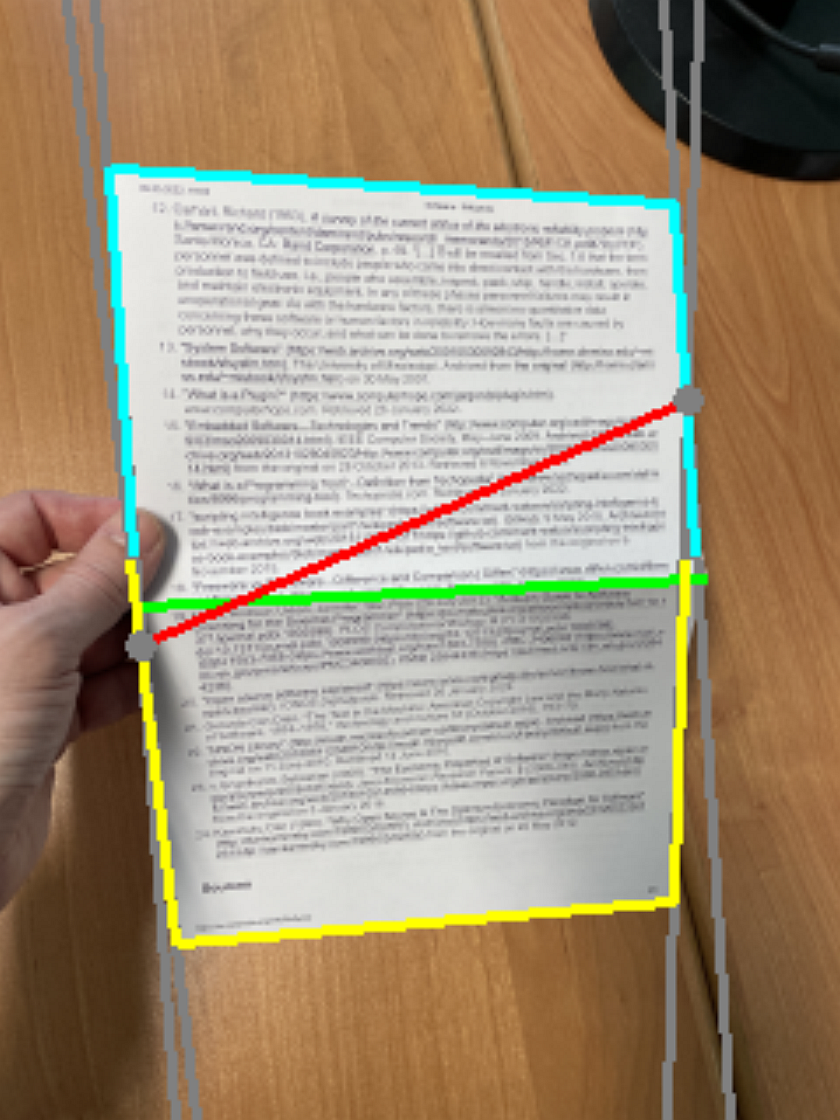}}
  \centerline{(c)}
\end{minipage}
\hfill
\begin{minipage}[b]{0.0645\linewidth}
  \centering
  \centerline{\includegraphics[width=\textwidth]{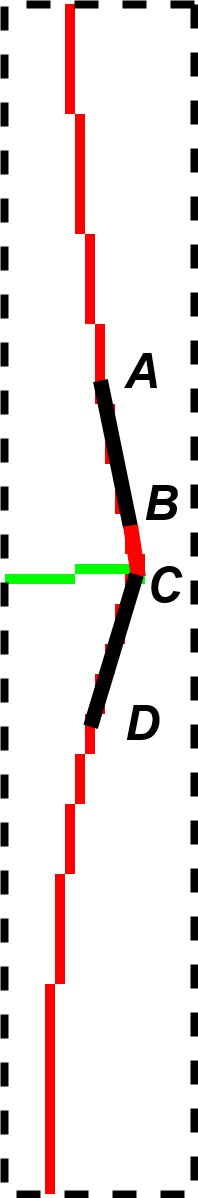}}
  \centerline{(d)}
\end{minipage}\\[1ex]
\begin{minipage}[b]{0.29\linewidth}
  \centering
  \centerline{\includegraphics[width=\textwidth]{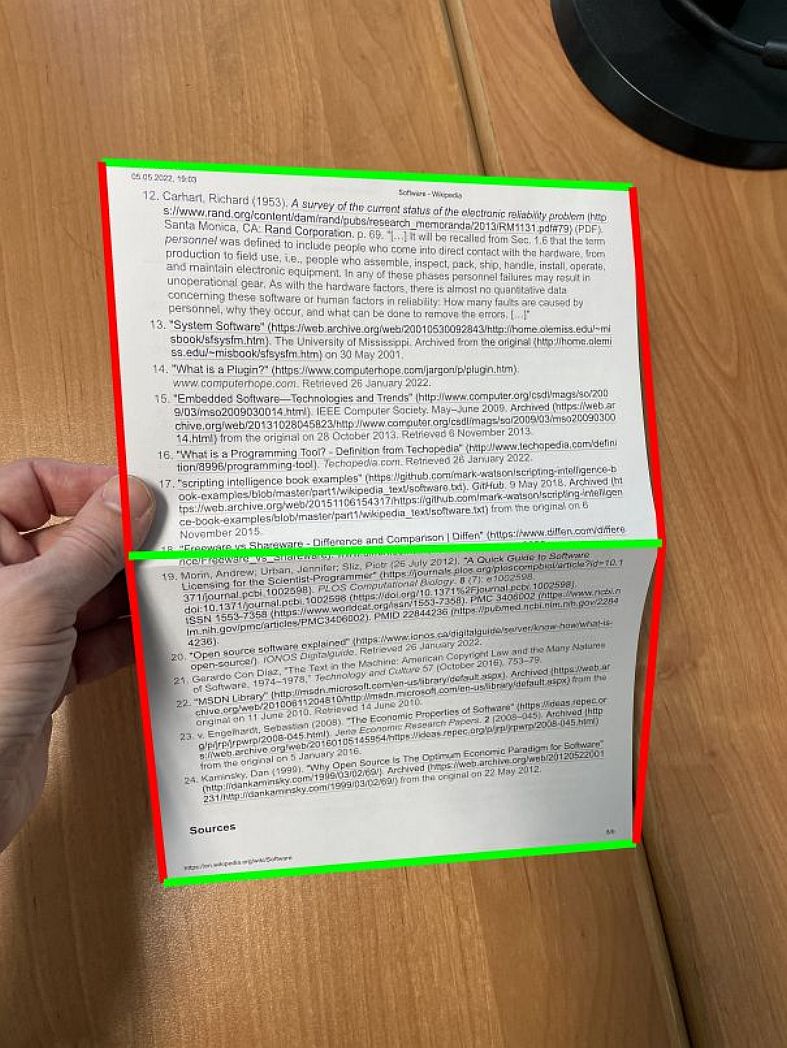}}
  \centerline{(e)}
\end{minipage}
\hfill
\begin{minipage}[b]{0.29\linewidth}
  \centering
  \centerline{\includegraphics[width=\textwidth]{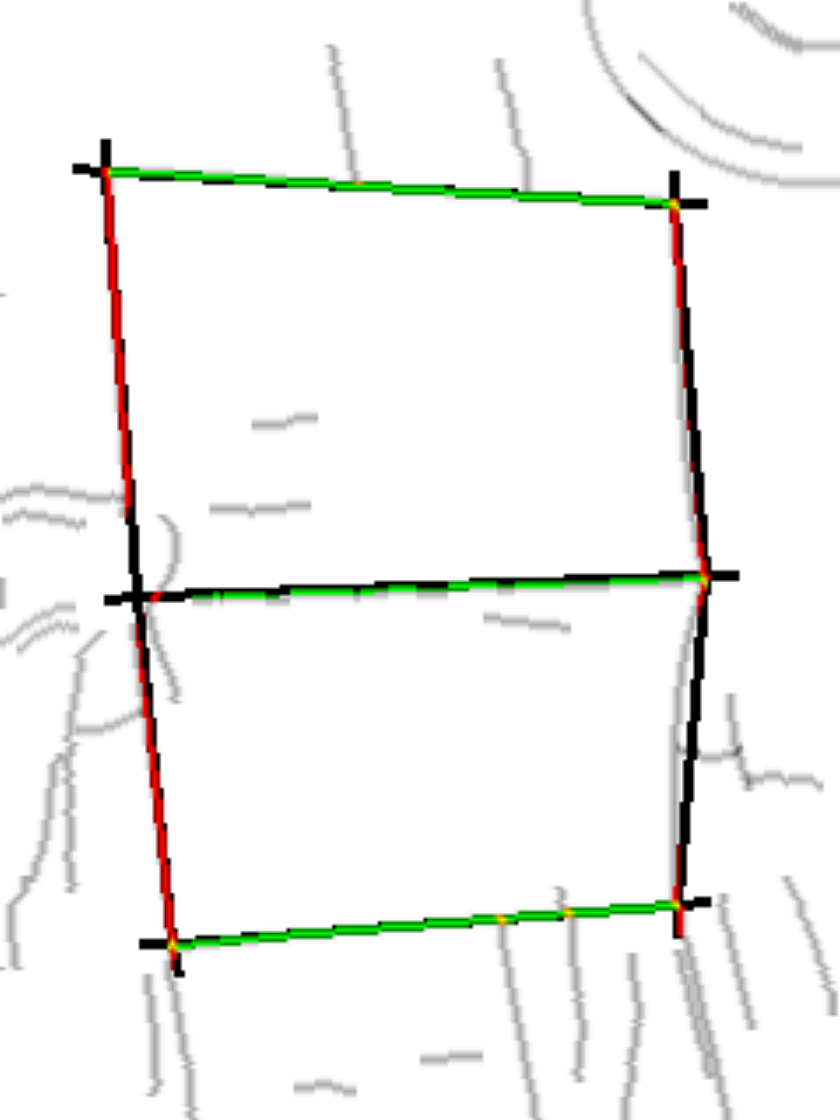}}
  \centerline{(f)}
\end{minipage}
\hfill
\begin{minipage}[b]{0.2707\linewidth}
  \centering
  \centerline{\includegraphics[width=\textwidth]{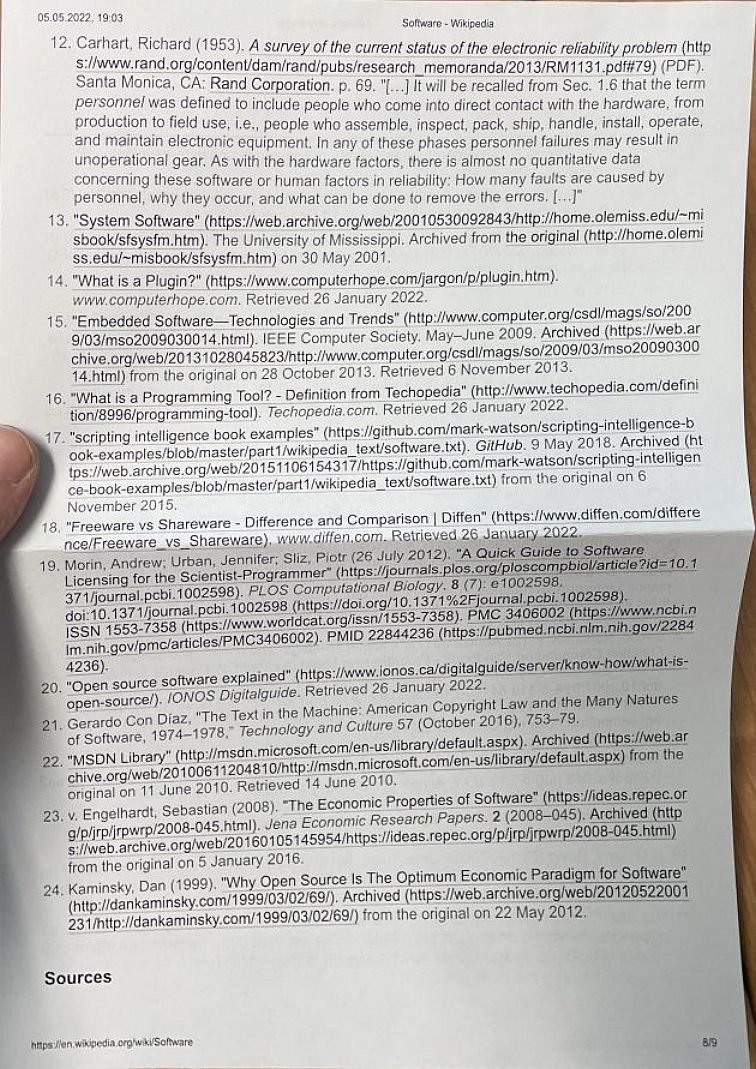}}
  \centerline{(g)}
\end{minipage}
\caption{(a) Horizontal (green) and vertical (red) edges on the ``merged'' edge map, (b)~detected lines (red and green) and the blue line splitting the image into halves, (c) a pair of quadrilaterals (the top one is depicted with blue, the bottom one with yellow), the true crease line (green) and crease line computed by vertical segments intersection (red), (d) a polyline $ABCD$ approximating a boundary fracture on the path graph marked by a dotted rectangle on (a), (e) detected hexangle, (f) edges lying along the detected hexangle, (g) rectified image.}
\label{fig:lines}
\end{figure}

In order to facilitate further understanding consider Fig.~\ref{fig:lines} with illustrations of Unfolder steps on a sample image. 
Fig.~\ref{fig:lines}.\textit{a} depicts the ``merged'' edge map: $E_v$ is shown with red color, $E_h$ -- with green.
The lines are illustrated on Fig.~\ref{fig:lines}.\textit{b}, $L_v$ is shown in red, $L_h$ is shown in green, the line separating the image into halves is shown in blue.

\begin{center}
	\underline{2.2 Hexangle formation}
\end{center}
Here and hereafter, by \textit{crease points} we mean the points at which the \textit{crease line} and the outer document edge intersect (see point $F$ on Fig.~\ref{fig:lines}.\textit{a}).

Let us consider a pair of quadrilaterals $q^t \in Q^t, \, q^b \in Q^b$.
To generate a hexangle all the segments of these quadrilaterals except from the ones lying on the line $l_s$ are analyzed.
One might suppose that the vertical segments of $q^t$ and $q^b$ can be intersected to get a hexangle.
However, since in general the halves of the document are not flat and their contours are not strictly line segments, the aforementioned intersection points of the segments may not correspond to \textit{crease points}.
Let us consider Fig.~\ref{fig:lines}.\textit{c}.
A cian polyline represents the quadrilateral $q^t$ and a yellow polyline represents the quadrilateral $q^b$, the line $l_s$ is not depicted for simplification.
If the vertical segments are simply intersected, the detected \textit{crease line} (depicted with red) will not correspond to the ``real'' one (depicted with green).

Thereby we analyze edges rather than the line segments.
Consider the edge map $E_v$.
It by-design [5] contains edges being non-zero pixels united into components $e_v$ of 8-connectivity having no more than one pixel in each image row.
These components hereinafter are referred to as \textit{path graphs}.
For every vertical segment of the quadrilaterals $q^t, q^b$, a corresponding \textit{path graph} from the edge map is distinguished and taken into consideration.


Namely, for every \textit{path graph} $e_v$ we calculate the number of its pixels at distance not farther than $\delta_{min}^{e}$ ($=3$) from a given segment $s$ 
(Note: here and hereafter all the manual parameters of the proposed algorithm are denoted with Greek letters, their values used in the experimental section are written in brackets.
The units used to measure metric distances coincide with the 1-pixel grid on the image $I$.
Most of these parameters were selected independently by analyzing the errors occurring in the subprograms of the present algorithm to reduce overfitting effects).
After that the \textit{path graph} with the highest such number is selected and is said to be corresponding to $s$:
\begin{equation}
e_v^{*}(s) = \underset{e_v \subset E_v}{\arg\,\max}\#\{p \in e_v : \text{dist}(p, s) < \delta_{min}^{e}\}
\end{equation}
If this number for $e_v^{*}$ is greater than $\rho_{min}$ ($=50\%$) of the length of $s$, then it is considered valid.

Once every segment has a \textit{path graph} assigned to it, we check if there is a pair of vertical segments $s^t \in q^t, s^b \in q^b$ such that $e_v^{*}(s^t) = e_v^{*}(s^b)$.
We shall denote such a \textit{path graph} $e_v^{**}$ and say that it is common to two vertical segments.
If there is such a pair on the left or on the right side, we generate up to four document location alternatives for the considered pair $\left(q^t, q^b\right)$.
We use two primitives to generate them: the \textit{angle} approximating the fracture of the boundary generated by the paper fold (see Sec.~2.2.1) and the \textit{crease line} (see Sec.~2.2.2).
The alternatives are defined by these primitives as follows: in $h_1$ -- by the two \textit{angles}, in $h_2$ -- by the left \textit{angle} and the \textit{crease line}, in $h_3$ -- by the right \textit{angle} and the \textit{crease line}, in $h_4$ -- only by the \textit{crease line}.


\begin{center}
	{2.2.1 The boundary fracture estimation}
\end{center}
Let us consider the \textit{path graph} $e_v^{**}$ common to the two vertical segments $s^t, s^b$ (for instance, the red \textit{path graph} inside the dotted region on Fig.~\ref{fig:lines}.\textit{a}).
The goal is to detect the \textit{crease point} ($F$ on the same figure) as a point with the maximal boundary fracture.

To estimate the \textit{crease point}, a 3-segment polyline $ABCD$ with a central segment $BC$ and outer segments $AB$ and $CD$ is searched along the \textit{path graph}. 
Let us index the points of $e_v^{**}$ from $0$ to $N := \# e_v^{**}$ top to bottom and denote the index of an arbitrary point $p$ of $e_v^{**}$ as $i(p)$.
Starting at each point of $e_v^{**}$, we shall construct an optimal polyline in the following way. 
Consider an arbitrary point of $e_v^{**}$ and suppose it is the point $B$ of $ABCD$.
First, given that $i(\overline{C}) > i(B)$, the shortest line segment $B\overline{C}^*$ is constructed so that the linear regression for points between them would have a quadratic error exceeding $\varepsilon_c (= 0.2)$ and take the point $C^{*}$ as the point before $\overline{C}^*$:
\begin{equation}
    \scriptsize
    C^{*} =  \underset{C \in e_v^{**}}{\arg\,\max} \left\{i(C) < i\left(\underset{\overline{C} \in e_v^{**}}{\arg\,\min}\left\{|B\overline{C}| : \sum\limits_{p \in e_v^{**}:\, i(B) < i(p) < i\left(\overline{C}\right)} \text{dist}_A\left(p, B\overline{C}\right)^2 \geqslant \varepsilon_c \right\}\right)\right\}.
\end{equation}
Here $\text{dist}_A\left(p, B\overline{C}\right)$ denotes the algebraic distance between a point and a line.

Then, given that $i(B) > i(A)$, a line segment $A^*B$ with the maximal possible length is constructed so that the same regression would allow for the quadratic error exceeding $\varepsilon_o (= 1)$. 
Under the same constraints, given that $i(C^*) < i(D)$, a line segment $C^*D^*$ is constructed (see Fig. \ref{fig:lines}.\textit{d}).
The \textit{angle} of the given polyline is introduced as the largest angle between lines containing segments $A^*B$ and $C^*D^*$.
The vertex of the \textit{angle} is the intersection of segments $A^*B$ and $C^*D^*$.
If these lines do not intersect, the midpoint of the segment $BC^*$ is considered to be its vertex.

After such a polyline is constructed for each point of the \textit{path graph}, the point with the smallest \textit{angle} of the corresponding polyline is selected.
If this \textit{angle} exceeds $\varphi_{max}^{c}$($=170$ degrees) it is assumed that the fracture of the path graph in question was not detected. 
Otherwise, the vertex of this \textit{angle} is the output of the \textit{crease point} detection algorithm.


\begin{center}
	{2.2.2 Crease line detection}
\end{center}
The alternative $h_2$ is based on the left \textit{crease point} and the \textit{crease line}, the alternative $h_3$ considers the right \textit{crease point} and the \textit{crease line}, and the alternative $h_4$ is generated by the \textit{crease line} only.
First, let us describe the \textit{crease line} detection algorithm for each alternative, and then we will define the remaining vertices.

The \textit{crease line} for $h_2$ and $h_3$ is selected from $L_h$ containing all the \textit{horizontal} lines.
The desired \textit{crease line} is chosen as the \textit{brightest} (in terms of the Hough image) among the lines non-further than $\delta_{max}^c (=15)$ from the \textit{crease point}.
If there are no lines in the neighborhood of the considered point, the alternatives $h_2, h_3$ are not considered further.

As for $h_4$, the \textit{crease line} is detected as follows.
First, we calculate the intersection points $P^t_l$, $P^t_r$, $P^b_l$ and $P^b_r$ of lines containing the \textit{horizontal} segments of $q^t$ and $q^b$ with the left and right image boundaries.
Then for each line from the set of \textit{horizontal} lines, we search for the intersections with these image boundaries.
Then we select the lines intersecting the left image boundary between $P^t_l$ and $P^b_l$ at a distance of at least $\delta_{min}^{b} (=10)$ from each, and the right boundary between $P^t_r$ and $P^b_r$ at the same distances.

From the selected lines, we choose the line with the largest value in the Hough image as the \textit{crease line}.
Then, we consider the \textit{crease line}'s \textit{path graphs} with a length of at least $\rho_{min}^{l, 1} (=40\%)$ of the image width.
The correspondence between the \textit{crease line} and a \textit{path graph} is determined identically as described previously in Sec.~2.2. 
For each of these \textit{path graphs}, we check whether or not at least $\rho_{min}^{l, 2} (=90\%)$ of its length lies within $q^t \cup q^b$.
If some \textit{path graph} does not pass the check, it is no longer considered.

Then we extend the remaining \textit{path graphs} by adding $\beta (=3)$ pixels to its left in the same image row with its leftmost point and $\beta$ pixels to its right to the right from its rightmost point.
After that we intersect it with the \textit{path graph} $e_v^{**}$.
If there are such intersections both on the left and on the right, we pick the left one.
This intersection point (if exists) is considered to be one of the vertices of the hexangle.

For the alternatives $h_2, h_3, h_4$ five vertices (all except one vertex on the \textit{crease line}) and the \textit{crease line} are known.
Without loss of generality, let the left vertex be known.
To determine the right vertex, we intersect the \textit{crease line} with the right vertical segments of $q^t$ and $q^b$.
The midpoint between these intersections is the last vertex of the hexangle.

\begin{center}
	\underline{2.3 Hexangle correction to fit the continuity criterion}
\end{center}

\begin{figure}[!ht]
    \centering
    \includegraphics[width=\textwidth]{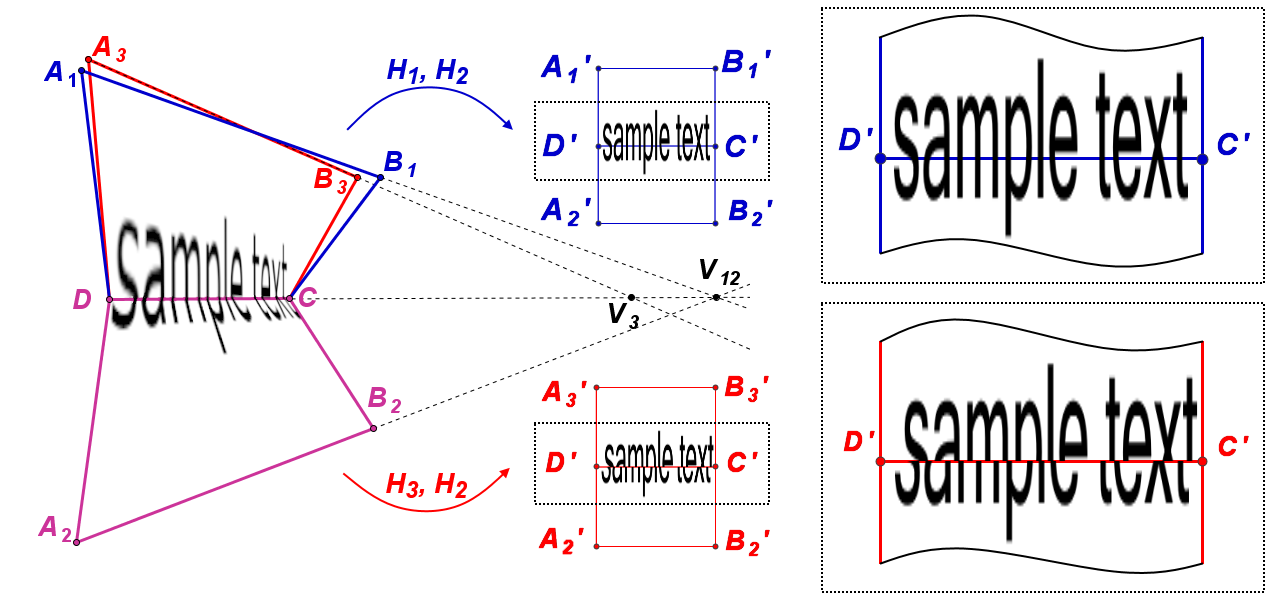}
    \caption{Examples of a continuous ($H_1$, $H_2$) and discontinuous ($H_3$, $H_2$) mapping.}
    \label{fig:two_planes}
\end{figure}

When using a mapping defined by two projective transformations to rectify the hexangle of the document's outer edges: the upper half of the document is projectively transformed into the upper part of the rectified image, the lower half of the document is projectively transformed into the lower part of the rectified image, there is a possibility that the text will be split at the junction of the two halves (see Fig.~\ref{fig:fragmentation}.\textit{c} and the red part of Fig.~\ref{fig:two_planes}).

The following criterion guarantees the continuity.
Let there be two quadrilaterals on the plane: $A_1B_1CD$ and $A_2B_2CD$ with the adjacent side $CD$, located in different half-planes relative to $CD$. 
Let there be rectangles $A_1'B_1'C'D'$ and $A_2'B_2'C'D'$ with the adjacent side $C'D'$ in different half-planes relative to it. 
Let $H_1$ be a projective transformation from $A_1B_1CD$ to $A_1'B_1'C'D'$, and $H_2$ be a projective transformation from $A_2B_2CD$ to $A_2'B_2'C'D'$.
Let $H$ be a mapping equal to $H_1$ on the half-plane with $A_1B_1CD$ and equal to $H_2$ on the half-plane with $A_2B_2CD$.
Then $H$ is correctly defined on the segment $CD$ if and only if the intersecting lines $A_1B_1, A_2B_2$ and $CD$ share a common point
(see the blue part of Fig.~\ref{fig:two_planes}). 
The proof of this statement is listed in Appendix A.

Let us transform each of the document location alternatives according to this criterion as follows.
For the tree \textit{horizontal} segments of the hexangle an algorithm estimating three lines passing through one point and approximating these segments~[42] is employed.
After acquiring these lines, the vertices of the hexangle are recalculated: the
resulting top and bottom lines are intersected with the corresponding vertical lines. 
The resulting \textit{crease line} is intersected with the top vertical segments.
This operator for reducing the hexagon to the model of a single horizontal vanishing point will be denoted $V$.
Thus, the hexagon $h$ is changed to a new one, $h' = V(h)$.

\begin{center}
	\underline{2.4 Hexangle filtration and ranking}
\end{center}

For each hexangle from the set $\{h\}$, we calculate its \textit{contour score} as follows.
The edge maps $E_v, E_h$ are subjected to the Gaussian smoothing with a variance of $\sigma (=1.83)$.
The \textit{contour score} considers seven segments: the sides of the hexangle and the \textit{crease segment}.
For every segment, the following values are calculated: the sum of pixels along the segment $p_m$ and the number of zero-pixels along the segment $r_m$.
The number of zero pixels along each segment is summed and divided by the total length of all segments $l$, resulting in the ratio $r = \frac{\sum r_m}{l}$.
We also calculate the penalty value $q$ as proposed in [43].
Consider a non-crease vertex $v_n$ of a hexangle.
The sum of $\beta_p (=10)$ pixels nearest to $v_n$ outside of $h$ along the sides intersecting in $v_n$ is added to $q$.
As for the crease vertex $v_c$, the sum of $\beta_p (=10)$ pixels nearest to $v_c$ outside of $h$ along the crease line is added to $q$.
Then the contour score of the hexagon is the value 
\begin{equation}\footnotesize
S(h) = \frac{\sum p_m}{r+1} - q.
\end{equation}


Let us consider Figures \ref{fig:lines}.\textit{e}, \ref{fig:lines}.\textit{f}. 
The first one illustrates the hexagon $h$ as a set of its segments.
The latter are considered in $S(h)$.
The second one shows the merged edge map with the regions considered in $S(h)$.
The color coding on this illustration is the following.
White- and gray-colored pixels are out of consideration.
Black-colored pixels represent the absence of an edge, the colored pixels represent its presence.


From the set of hexagons $\{h'\}$, we select $h'^{*}$ such that
\begin{equation}\footnotesize
h'^{*} = \underset{h'}{\arg\,\max}\left\{S\left(V^{-1}\left(h'\right)\right) : \max\left\{\left|R_t\left(h'\right) - R_0\right|, \left|R_b\left(h'\right) - R_0\right|\right\} < 0.3 R_0 \right\},    
\end{equation}
where $V^{-1}(h')$ is the hexagon $h$ from which $h'$ was obtained by the operator~$V$, $R_t(h')$ and $R_b(h')$ are the aspect ratios [44] for the upper and bottom half of $h^{*}$ in the pinhole camera model with the focal length of $\lambda  (=0.705)$ and with the principle point in the center of the image, $R_0$ is the true aspect ratio of $\frac{297}{210}$ for A4 paper.

We calculate the distances between corresponding vertices of $h'^{*}$ and $h^{*} = V^{-1}(h'^{*})$ as well as the angles between new and old horizontal segments.
If any of the angles exceeds $\varphi_{max}^{V}$ (=2.56 degrees), or any of the vertices deviations is greater than $\rho_{max}^{V} (=1\%)$ of the image height, then we assume that such correction may corrupt the horizontal text lines and this hexangle is rejected and the output of the algorithm is trivial (no hexangle). 

\begin{figure}[t]
\begin{minipage}[b]{0.23\linewidth}
  \centering
  \centerline{\includegraphics[width=\textwidth]{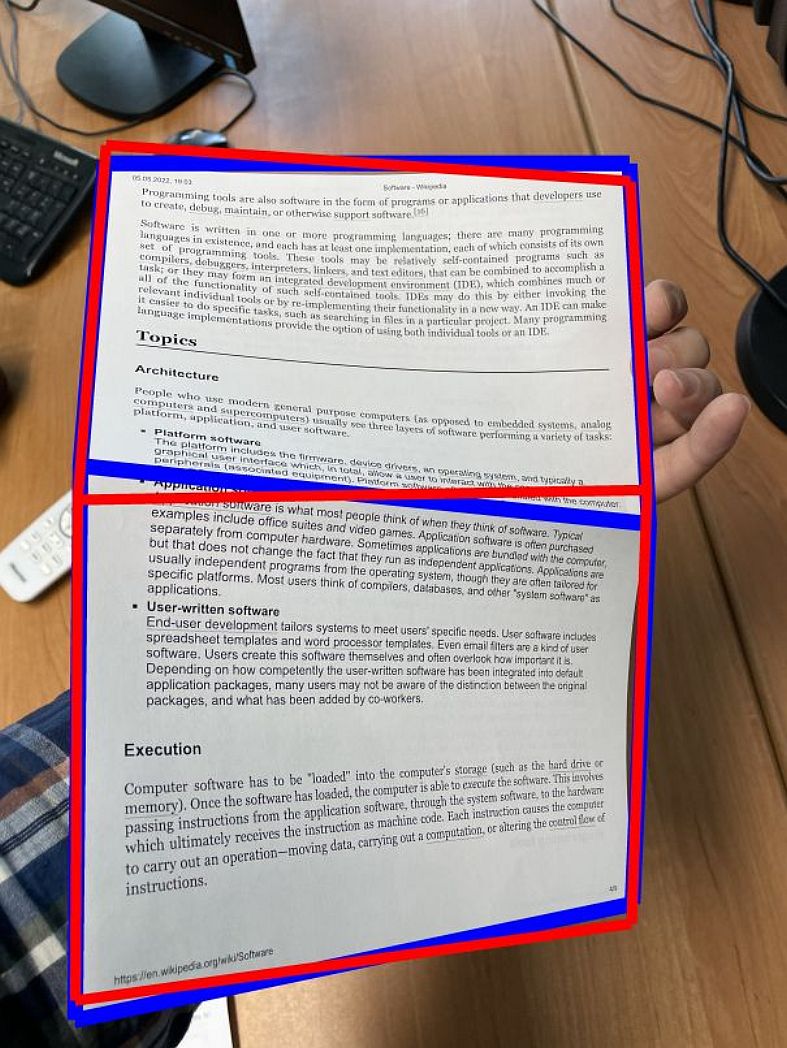}}
  \centerline{(a) }
\end{minipage}
\hfill
\begin{minipage}[b]{0.217\linewidth}
  \centering
  \centerline{\includegraphics[width=\textwidth]{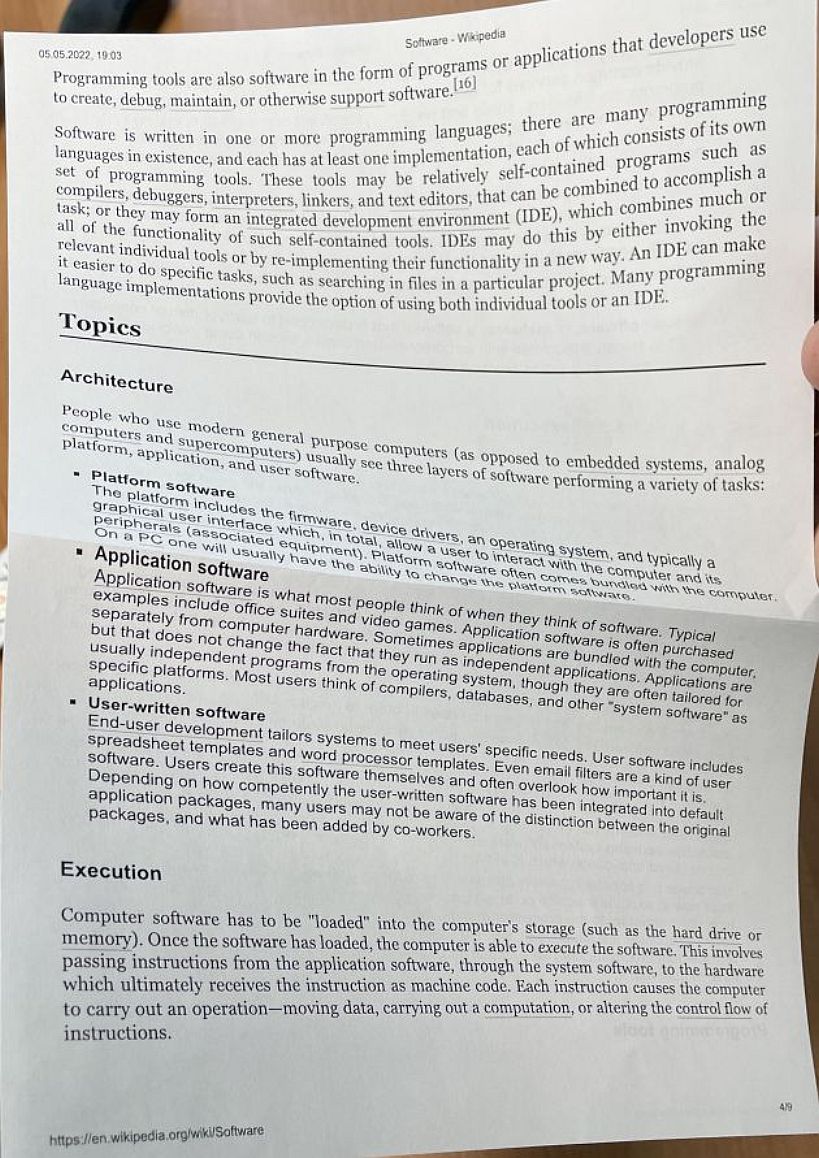}}
  \centerline{(b)}
\end{minipage}
\hfill
\begin{minipage}[b]{0.23\linewidth}
  \centering
  \centerline{\includegraphics[width=\textwidth]{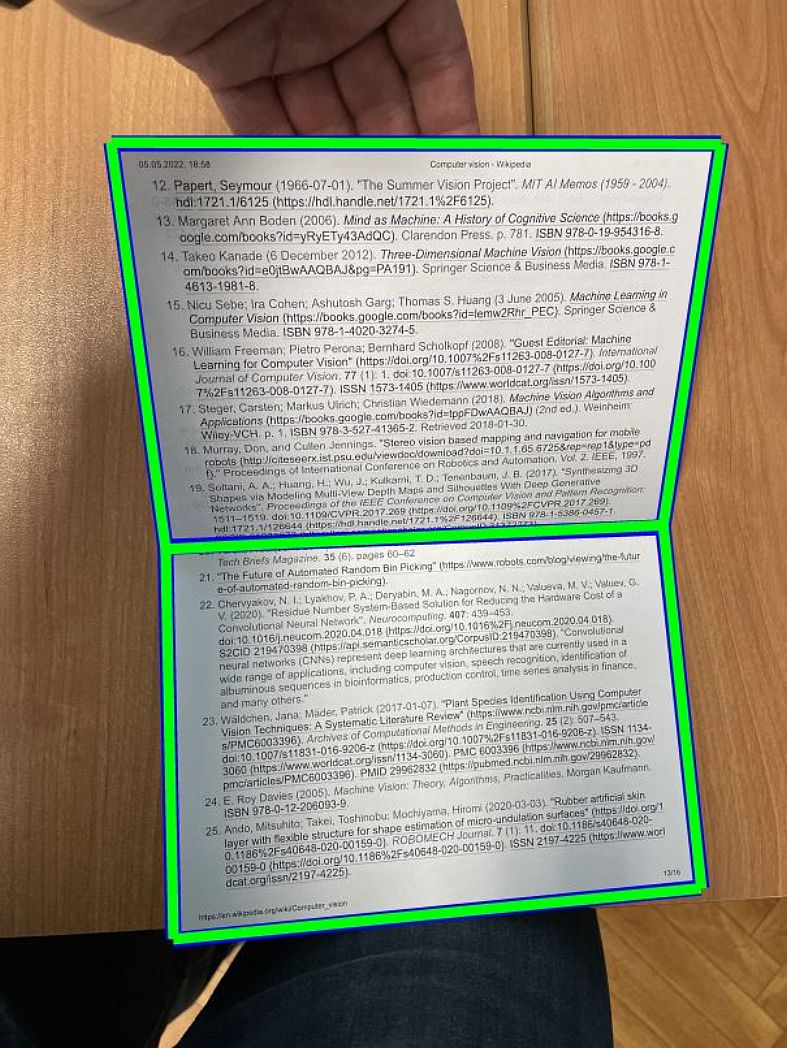}}
  \centerline{(c)}
\end{minipage}
\hfill
\begin{minipage}[b]{0.217\linewidth}
  \centering
  \centerline{\includegraphics[width=\textwidth]{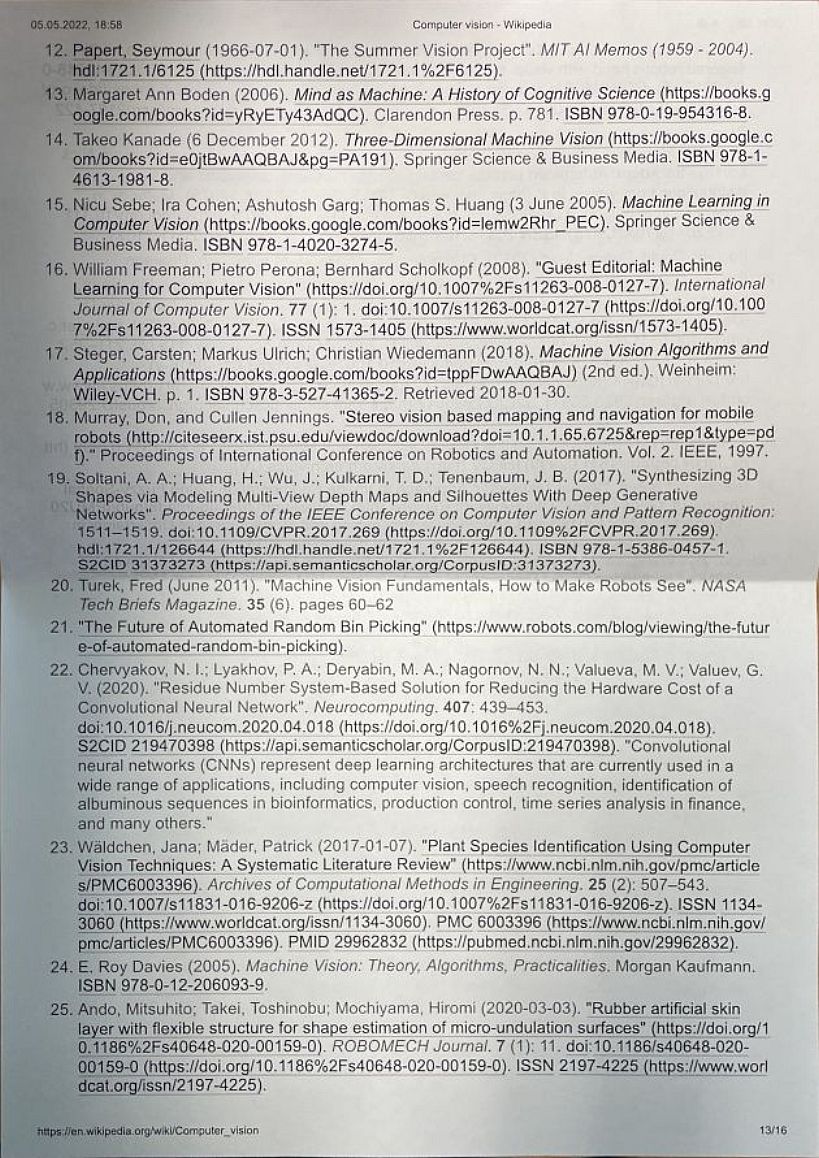}}
  \centerline{(d)}
\end{minipage}
\caption{Examples of a rejected and accepted rectification: (a)~the input sample image with a detected hexangle (blue) and its correction (red), which is rejected, (b)~the rectification of this image by the red hexangle, (c)~the input sample image with a detected hexangle (blue) and its correction (green), which is accepted, (d) the rectification of this image by the green hexangle.
}
\label{fig:helihexa}
\end{figure}

Let us consider Fig.~\ref{fig:helihexa}.
It illustrates two different images.
In both cases the formed hexangles (depicted with blue in Fig.~\ref{fig:helihexa}.\textit{a, c}) correspond to actual document position.
The document on the first image does not fit the model of two planes, so the hexangle $h^{*}$ is heavily affected by the operator $V$ and the rectification by the corrected hexangle $h'^{*}$ (depicted with red in Fig.~\ref{fig:helihexa}.\textit{a}) is corrupted  (Fig.~\ref{fig:helihexa}.\textit{b}).
The document on the second image fits the model, so the corresponding corrected hexangle $h'^{*}$ (depicted with green in Fig.~\ref{fig:helihexa}.\textit{c}) is accepted and the rectification (Fig.~\ref{fig:helihexa}.\textit{d}) is fine.



\begin{center}
	\underline{2.5 Projective transformation}
\end{center}
If the selected hexangle $h'^{*}$ is rejected, the original image is returned as an output of Unfolder.
Otherwise, each half of the hexangle is transformed projectively (using bilinear interpolation as described in [45]) onto the corresponding half of the rectified image ($2100 \times 2970$).
This image is the output of the Unfolder algorithm (see Fig.~\ref{fig:lines}.\textit{g}).


\begin{center}
	{\bf 3. Dataset and performance analysis}
\end{center}

\begin{figure}[ht!]
\begin{minipage}[b]{0.23\linewidth}
  \centering
  \centerline{\includegraphics[width=\textwidth]{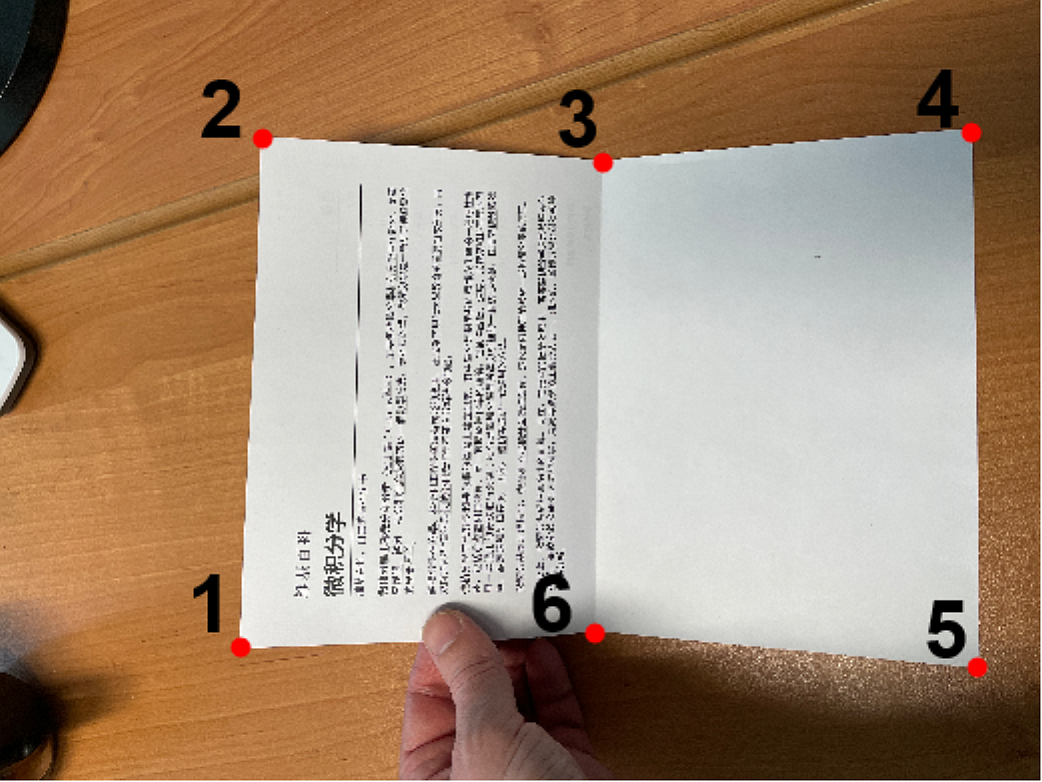}}
  \centerline{(a) }
\end{minipage}
\hfill
\begin{minipage}[b]{0.23\linewidth}
  \centering
  \centerline{\includegraphics[width=\textwidth]{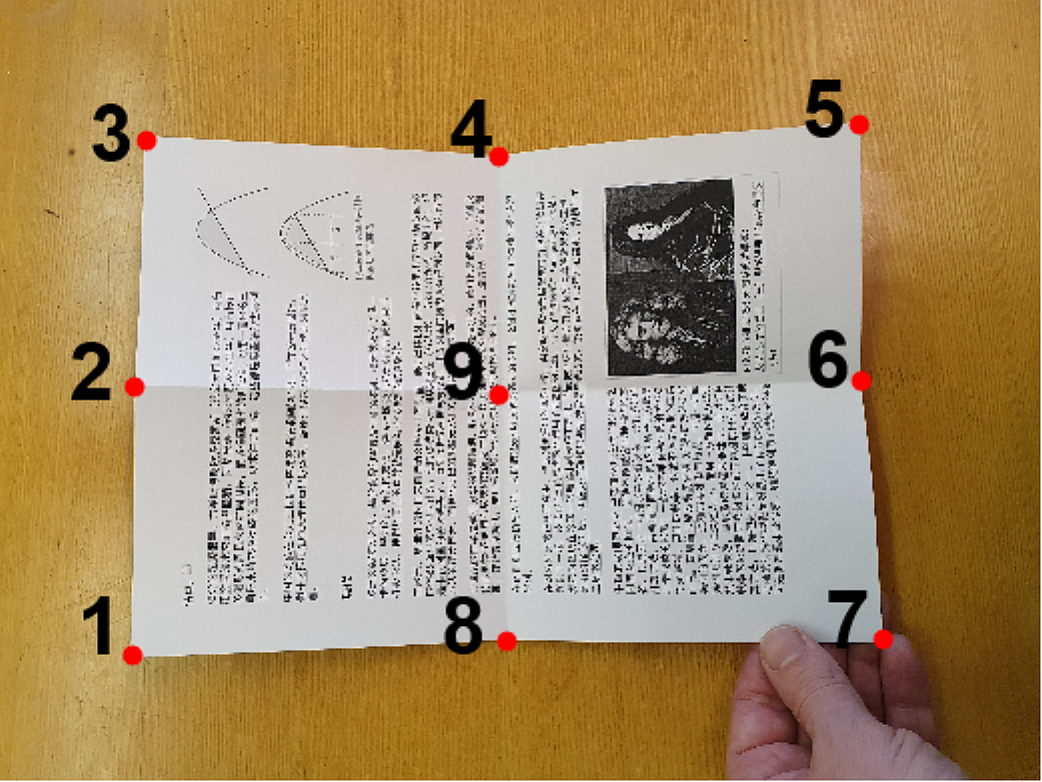}}
  \centerline{(b)}
\end{minipage}
\hfill
\begin{minipage}[b]{0.23\linewidth}
  \centering
  \centerline{\includegraphics[width=\textwidth]{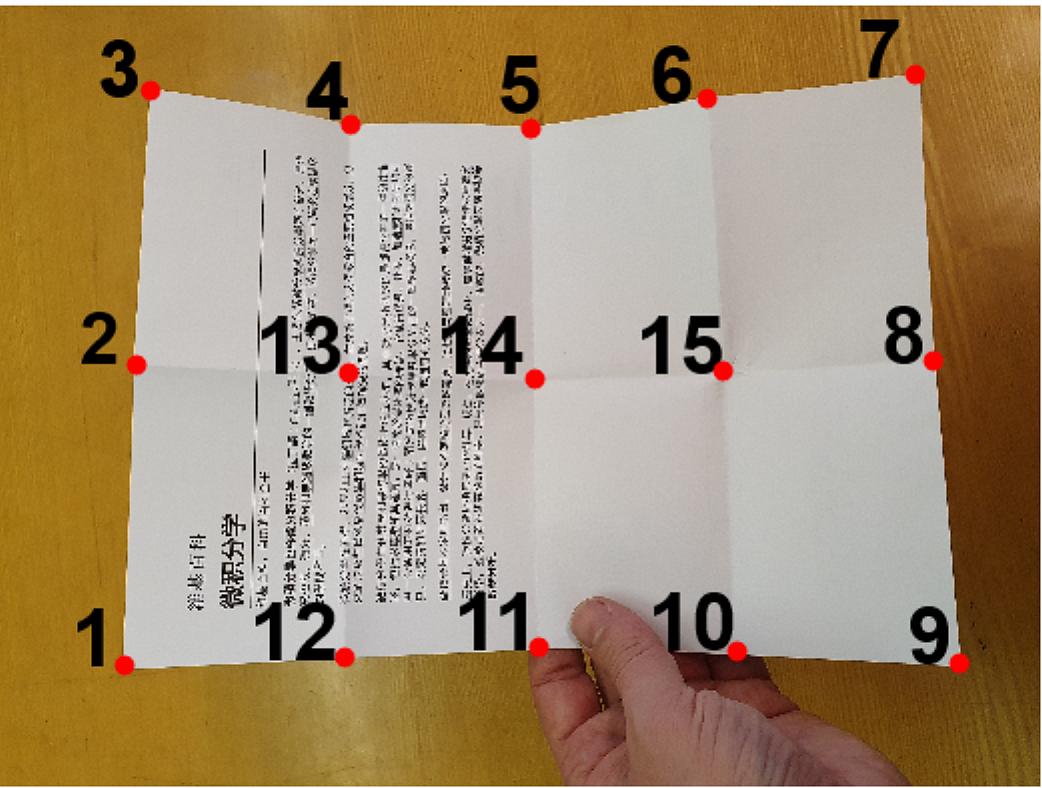}}
  \centerline{(c)}
\end{minipage}
\hfill
\begin{minipage}[b]{0.23\linewidth}
  \centering
  \centerline{\includegraphics[width=\textwidth]{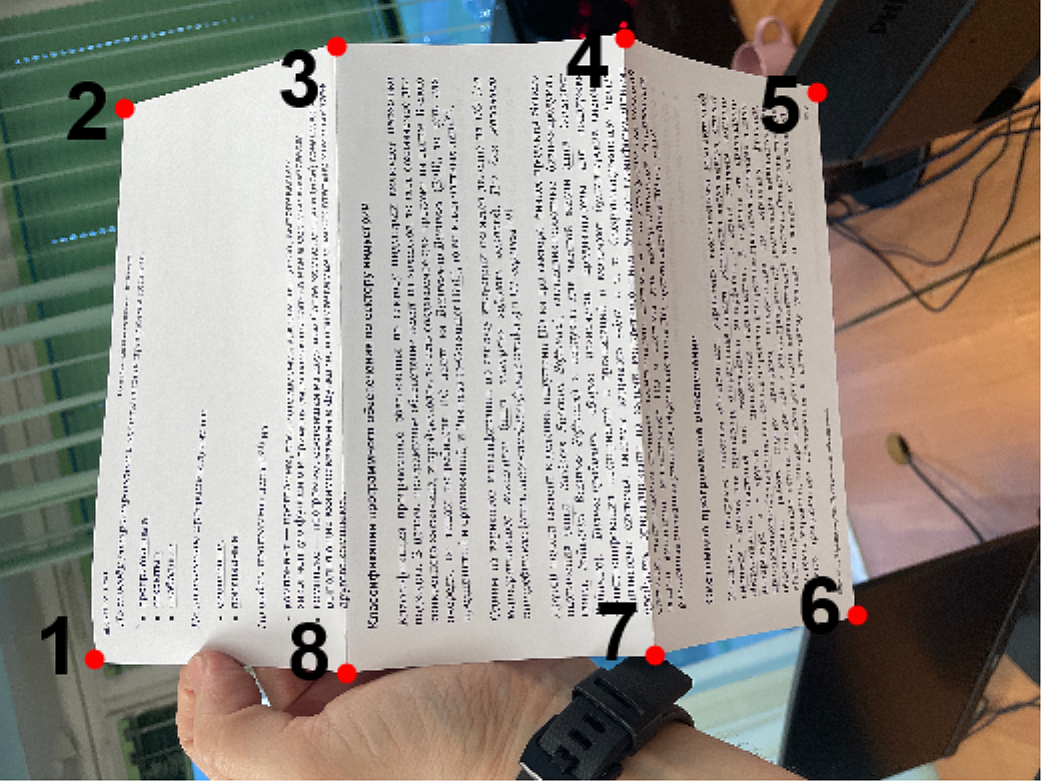}}
  \centerline{(d)}
\end{minipage}
\caption{Examples of vertex indexing in the annotation for the subsets (a) \textit{2fold}, (b)~\textit{4fold}, (c) \textit{8fold},  (d) \textit{3fold}.}
\label{fig:dataset_examples}
\end{figure}

\begin{center}
	\underline{3.1 Folded Document Images dataset}
\end{center}
We propose a brand new Folded Document Images (FDI) dataset which includes 1600 annotated A4 document images and can be accessed at
\url{ftp://smartengines.com/fdi}.
To create the dataset we selected 200 documents containing Wikipedia articles in Arabic, Chinese, English, Hindi, and Russian (40 documents in each language).
Using a laser printer, we printed out all the documents in two copies.
Both copies were used for creation the dataset with 4 different folding types (see Fig.~\ref{fig:dataset_examples}): (a) in half along the short edge -- hereinafter referred to as \textit{2fold}; (b) in half along the short edge, then in half along the long edge -- \textit{4fold}; (c) in half along the short edge twice and in half along the long edge -- \textit{8fold}; (d) along the short edge into thirds (as when folding into an envelope) -- \textit{3fold}.
The first copy was used for (a)-(c), the second for (d).
All images are captured with a smartphone (iPhone~12 or Samsung Galaxy S10) camera in 4032$\times$3024 resolution.
We captured 2 scenes for each document: the document is placed on the \textit{table}; the document is held in \textit{hand} and shot in front of the office furniture. 
In total, we captured $40 \times 5 \times 4 \times 2 = 1600$ images.

There are three folders in the root directory of the dataset: \textit{images}, \textit{annotation} and \textit{reference}.
In the first two directories all the files are ordered by folding type and then according to the captured scene.
The 1600 images in the \textit{images} folder are named using the following pattern: [$xt$]\_[$nnn$].\textit{jpg}, where $x$ is the number of folds, $t$ is the scene type (\textit{h} stands for \textit{hand}, \textit{t} stands for \textit{table}), $nnn$ is the ordinal number of the document. 
For instance, $2h\_001.jpg$ means the twofold document with number $001$ being held in hand on the captured image.
The reference images of the documents can be found in the {\it reference} folder: 200 files with pattern [$nnn$].\textit{pdf} in \textit{reference/pdf} folder and 200 files with pattern [$nnn$].\textit{tiff} in \textit{reference/tiff} folder. 
The PDF files were used for printing and then were converted to TIFF images which were employed to evaluate the accuracy of rectification. 
For each image [$xt$]\_[$nnn$].\textit{jpg}, the {\it annotation} folder contains the corresponding annotation  [$xt$]\_[$nnn$].\textit{jpg.json}.
Each JSON annotation file provides information on: the corresponding reference image (key \textit{reference}); the language of the document (\textit{language}); the folding type (\textit{folding}); the scene type (\textit{scene}); the reference coordinates of the points (\textit{vertices}) where the creases and/or edges of the document intersect.
The vertices are indexed starting from the upper left corner of the document spiraling clockwise (see Fig.~\ref{fig:dataset_examples}).
The coordinates of the vertices were determined manually via the VGG Image Annotator 2.0.11 [46].
In addition to above mentioned files each terminal directory contains \textit{runlist.lst} file, which specifies a list of files in the folder.

\begin{center}
	\underline{3.2 Metrics and baseline algorithms}
\end{center}
In 2018, K.~Ma \textit{et al.} proposed the DocUNet dataset [7] which include annotated images of $65$ arbitrarily folded (warped) documents.
The standard accuracy metrics for this dataset include the following statistics: multi-scale structural similarity index (MS-SSIM) [47]; local distortion (LD) [11] based on SIFT flow [48]; the Levenshtein distance or edit distance (ED) [49] between the optical character recognition results produced with Tesseract 5.0.1~[50] for a reference image and a rectified image; Character Error Rate (CER) calculated as the ratio of Levenshtein distance to the length of text recognized in the reference image.
In our experiments we will also use these metrics for performance analysis.
We execute Tesseract with all 5 languages enabled.


We chose DewarpNet~[24] and DocTr~[26] as a baseline for rectification of the FDI dataset.
DocTr is the current state-of-the-art according to the DocUNet benchmark, DewarpNet is one of the first neural network publications dedicated the document rectification problem.  
To execute these methods we used source code together with the pretrained models available in DocTr (\url{github.com/fh2019ustc/DocTr}, acquired on 16 Sep. 2022) and DewarpNet (\url{github.com/cvlab-stonybrook/DewarpNet}, acquired on 12 Oct. 2022) public repositories.
For our experiments the geometrical part of DocTr was used, annotated below as GeoTr.

\begin{center}
	\underline{3.3 Rectification accuracy on FDI}
\end{center}
Since Unfolder is designed under the assumption that the document has been folded in half along the short side, we will analyze in detail only the subset \textit{2fold}.
The performance of the proposed and baseline algorithms on it is presented in rows 1-6 of the Table~\ref{table:multibends}.

Row 1 contains the statistics of raw FDI images: SS (equals to 1~$-$ MS~SSIM), LD, CER, ED.
They can be referenced as the initial accuracy on images with no rectification algorithm applied.

The accuracy for \textbf{Unfolder} algorithm (see Sec.~2) is shown in row 2.

\textbf{GeoTr} and \textbf{DewarpNet} are tested in 2 setups each: on \textit{raw} FDI images (rows~3, 4) and on FDI images obtained by cropping over the document reference hexagon parsed from the annotation (rows~5, 6).
We will denote this additional crop step as C.
The crop is performed as follows: the maximum and minimum values for each of manually annotated polygon' coordinates are calculated.
The orthotropic rectangle enclosing the outermost points is expanded by 20 pixels in each direction, if it is possible without exceeding the image boundaries.
The image region corresponding to the extended orthotropic rectangle is the result of the crop step.


\begin{table*}[t]
\centering
\caption{Rectification accuracy on {\it 2fold}, {\it 4fold}, {\it 8fold}, {\it 3fold} subsets of FDI.}
{\footnotesize
\begin{tabularx}{\textwidth}{|c|c|Y|Y|Y|Y|Y|Y|Y|Y|}
\hline
\multirow[c]{2}{*}[0in]{\rotatebox{90}{Fold}} &\multirow{2}{*}{\backslashbox{System}{Accuracy}} & \multicolumn{4}{c}{Subset {\it hand}} & \multicolumn{4}{|c|}{Subset \textit{table}}
\\
\cline{3-10}
& & SS$\downarrow$ & LD$\downarrow$ & CER$\downarrow$ & ED$\downarrow$ & SS$\downarrow$ & LD$\downarrow$ & CER$\downarrow$ & ED$\downarrow$
\\
\hline

\multirow[c]{6}{*}[0in]{\rotatebox{90}{\textit{2fold}}}
& No algo (raw FDI) 	& 0.78 		&  51.24 		& 0.55 & 1814 & 0.75 & 46.39 & 0.43 & 1379 \\ \cline{2-10}
& Unfolder (our) & 0.59 &  15.20 & \textbf{0.35} & \textbf{1158}	& 0.54 & \textbf{7.34} &\textbf{0.30}& \textbf{941} \\ \cline{2-10}
& GeoTr &  0.61 & 21.72 & 0.51 & 1599 & \textbf{0.52} & 8.61 & 0.36 & 1133 \\ \cline{2-10}
& DewarpNet &  0.68 & 26.78 & 0.59 & 1893 & 0.69 & 22.10 & 0.55 & 1731 \\ \cline{2-10}
& С + GeoTr 	& \textbf{0.54} & \textbf{11.01} & 0.42 & 1319 & 0.55 & 10.04 & 0.49 & 1541 \\ \cline{2-10}
& С + DewarpNet 	& 0.58 & 11.26 & 0.39 & 1252 & 0.57 & 10.22 & 0.45 & 1373 \\ \hline\hline

\multirow[c]{6}{*}[0in]{\rotatebox{90}{\textit{4fold}}}
& No algo (raw FDI) & 0.72 & 51.53 & 0.50 & 1552 & 0.72 & 53.57 & 0.44 & 1402
\\
\cline{2-10}
& Unfolder (our) & 0.60 & 25.53 & 0.42 & 1338 & \textbf{0.53} & 11.60 & \textbf{0.36} & \textbf{1142} 
\\
\cline{2-10}
& GeoTr & 0.59 & 20.93 & 0.47 & 1453 & 0.55 & \textbf{10.45} & 0.42 & 1309
\\
\cline{2-10}
& DewarpNet & 0.67 & 25.14 & 0.55 & 1707 & 0.70 & 29.84 & 0.57 & 1742
\\
\cline{2-10}
& С + GeoTr & \textbf{0.54} & \textbf{9.69} & 0.45 & 1388 & 0.53 & 10.48 & 0.42 & 1316
\\
\cline{2-10}
& С + DewarpNet  & 0.58 & 11.09 & \textbf{0.41} & \textbf{1243} & 0.59 & 10.97 & 0.39 & 1197
\\
\hline
\hline
\multirow[c]{6}{*}[0in]{\rotatebox{90}{\textit{8fold}}}
& No algo (raw FDI) & 0.72 & 48.47 & 0.49 & 1557 & 0.72 & 51.08 & 0.48 & 1546
\\
\cline{2-10}
& Unfolder (our) & 0.69 & 40.28 & 0.46 & 1466 & 0.65 & 31.57 & 0.47 & 1517
\\
\cline{2-10}
& GeoTr & 0.60 & 20.73 & 0.47 & 1487 & \textbf{0.52} & \textbf{8.99} & 0.41 & 1309
\\
\cline{2-10}
& DewarpNet & 0.68 & 21.98 & 0.57 & 1756 & 0.70 & 27.07 & 0.58 & 1811
\\
\cline{2-10}
& С + GeoTr & \textbf{0.55} & \textbf{10.24} & 0.46 & 1442 & 0.53 & 9.88 & 0.44 & 1354
\\
\cline{2-10}
& С + DewarpNet  & 0.59 & 11.41 & \textbf{0.42} & \textbf{1298} & 0.57 & 10.59 & \textbf{0.39} & \textbf{1214}
\\
\hline
\hline
\multirow[c]{6}{*}[0in]{\rotatebox{90}{\textit{3fold}}}
& No algo (raw FDI) & 0.75 & 49.33 & 0.50 & 1574 & 0.72 & 52.58 & \textbf{0.33} & 1022
\\
\cline{2-10}
& Unfolder (our) & 0.73 & 48.21 & 0.49 & 1540 & 0.72 & 51.71 & \textbf{0.33} & \textbf{991}
\\
\cline{2-10}
& GeoTr & 0.63 & 24.85 & 0.54 & 1679 & \textbf{0.54} & 11.51 & 0.34 & 1050
\\
\cline{2-10}
& DewarpNet & 0.68 & 22.49 & 0.59 & 1822 & 0.70 & 29.84 & 0.57 & 1738 
\\
\cline{2-10}
& С + GeoTr & \textbf{0.56} & \textbf{11.70} & 0.45 & 1398 & \textbf{0.54} & \textbf{10.76} & 0.41 & 1294
\\
\cline{2-10}
& С + DewarpNet  & 0.58 & 12.47 & \textbf{0.39} & \textbf{1202} & 0.59 & 12.31 & 0.36 & 1094
\\
\hline
\end{tabularx}
}
\label{table:multibends}
\end{table*}

\begin{figure}[!ht]
\centering
\begin{minipage}[b]{0.191\linewidth}
  \centering
  \centerline{\includegraphics[width=\textwidth]{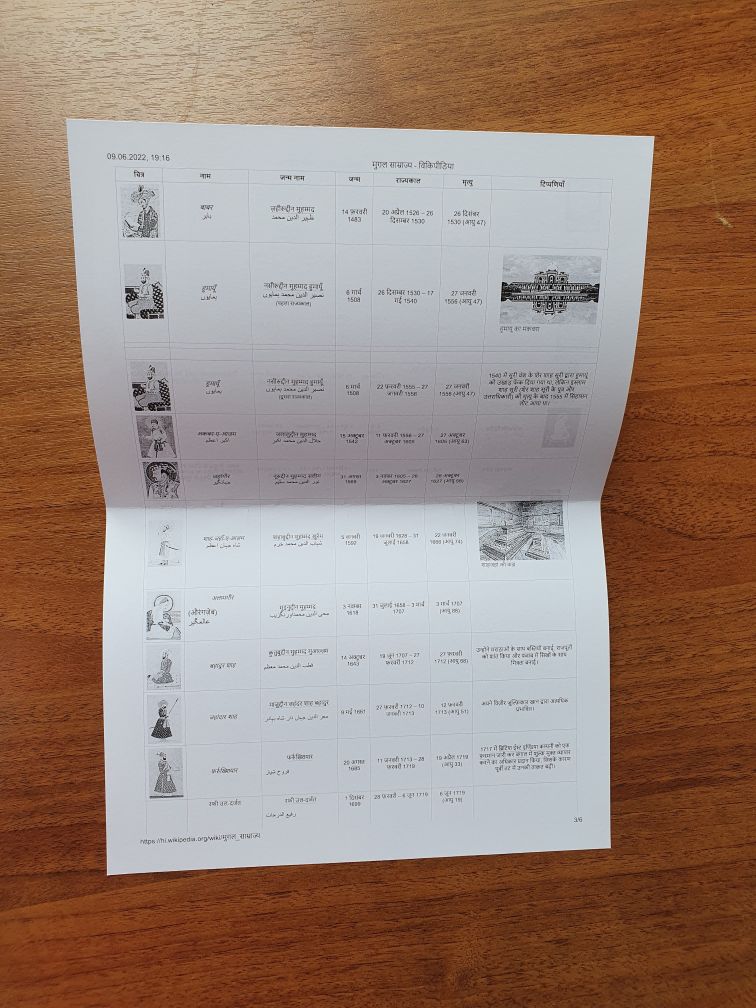}}
  \centerline{(a) }
\end{minipage}
\hfill
\begin{minipage}[b]{0.191\linewidth}
  \centering
  \centerline{\includegraphics[width=\textwidth]{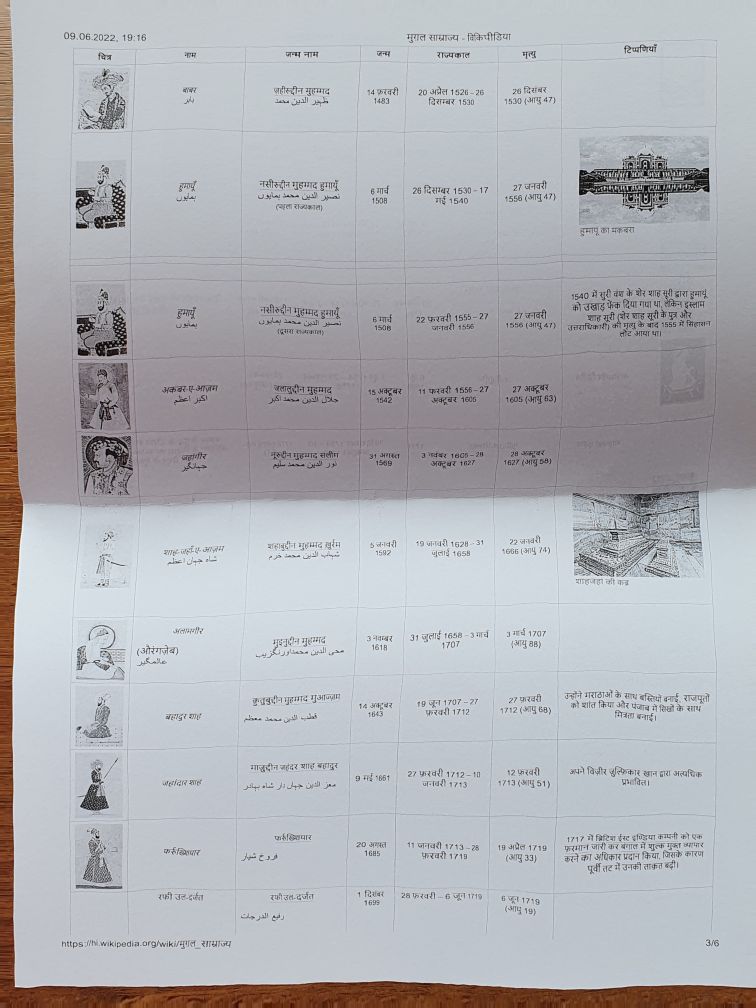}}
  \centerline{(b)}
\end{minipage}
\hfill
\begin{minipage}[b]{0.191\linewidth}
  \centering
  \centerline{\includegraphics[width=\textwidth]{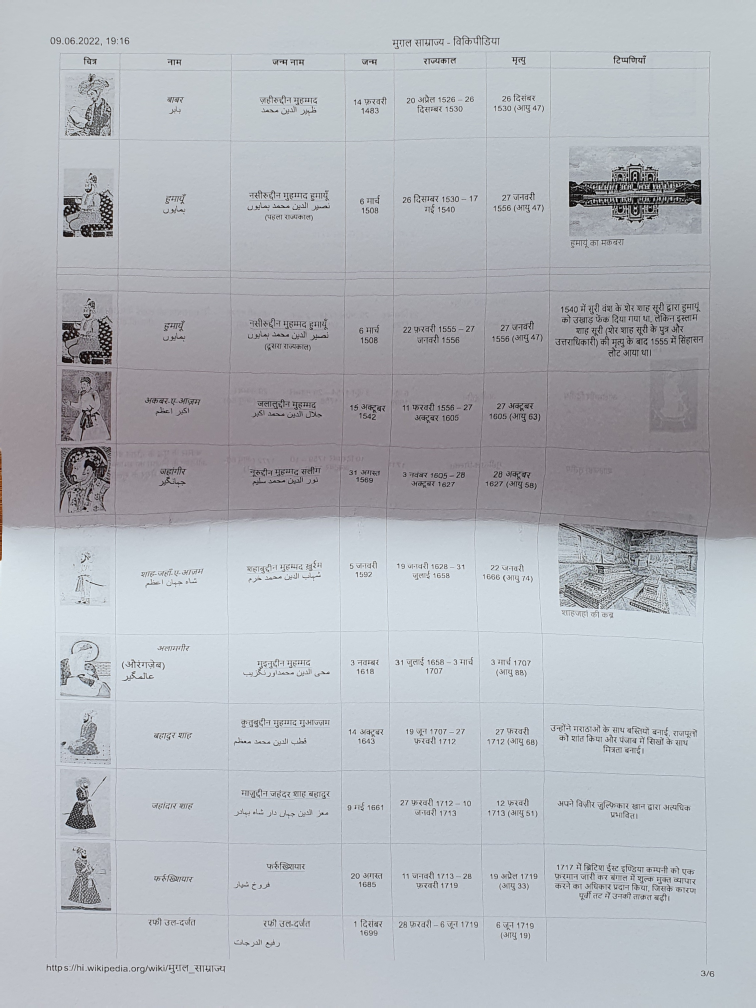}}
  \centerline{(c)}
\end{minipage}
\hfill
\begin{minipage}[b]{0.18\linewidth}
  \centering
  \centerline{\includegraphics[width=\textwidth]{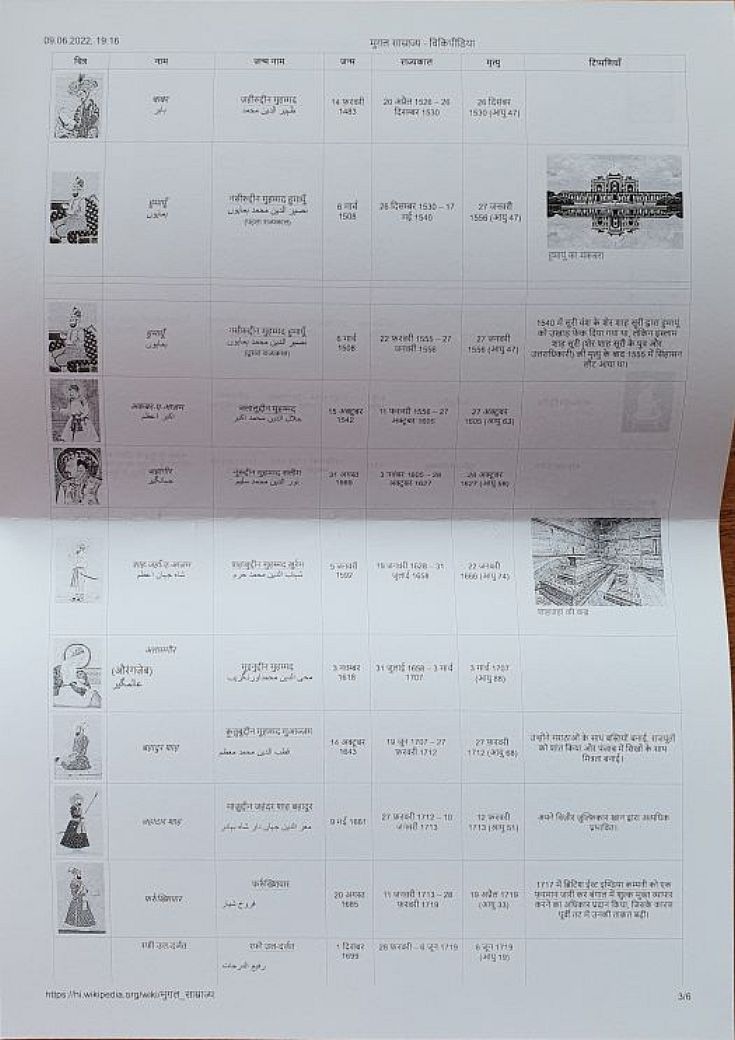}}
  \centerline{(d)}
\end{minipage}
\hfill
\begin{minipage}[b]{0.18\linewidth}
  \centering
  \centerline{\includegraphics[width=\textwidth]{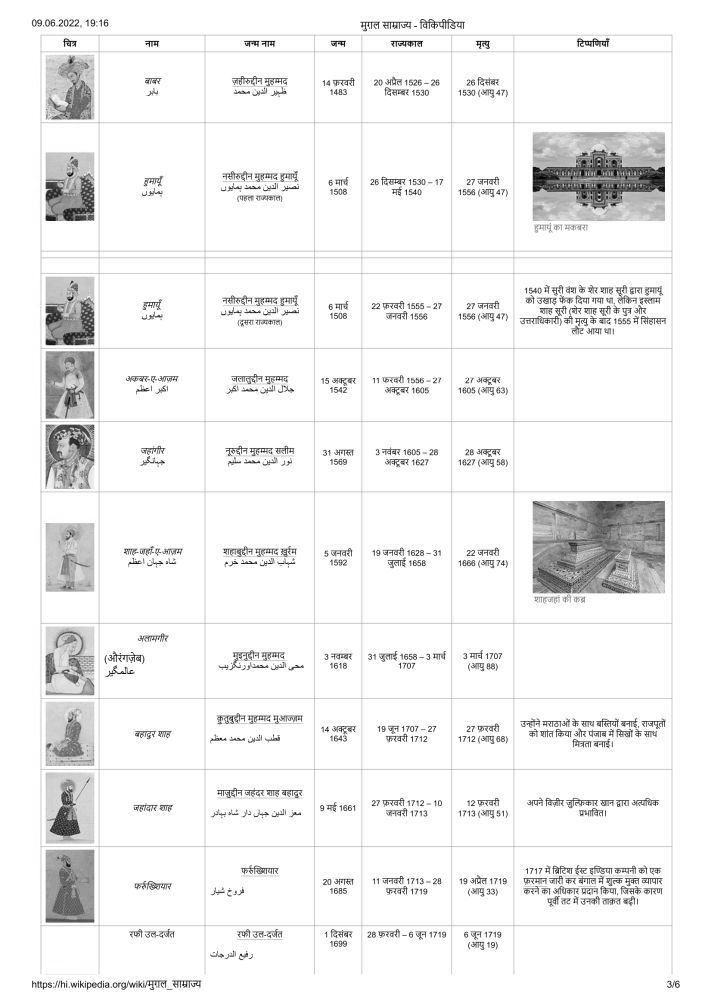}}
  \centerline{(e)}
\end{minipage}\\[1ex]
\begin{minipage}[b]{0.191\linewidth}
  \centering
  \centerline{\includegraphics[width=\textwidth]{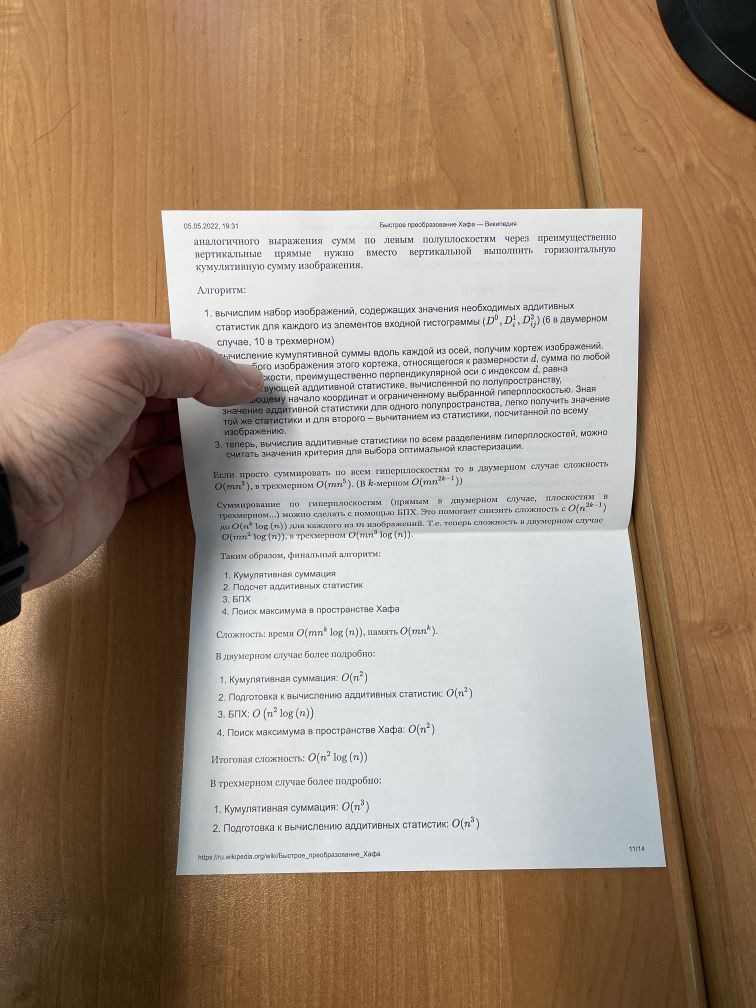}}
  \centerline{(f) }
\end{minipage}
\hfill
\begin{minipage}[b]{0.191\linewidth}
  \centering
  \centerline{\includegraphics[width=\textwidth]{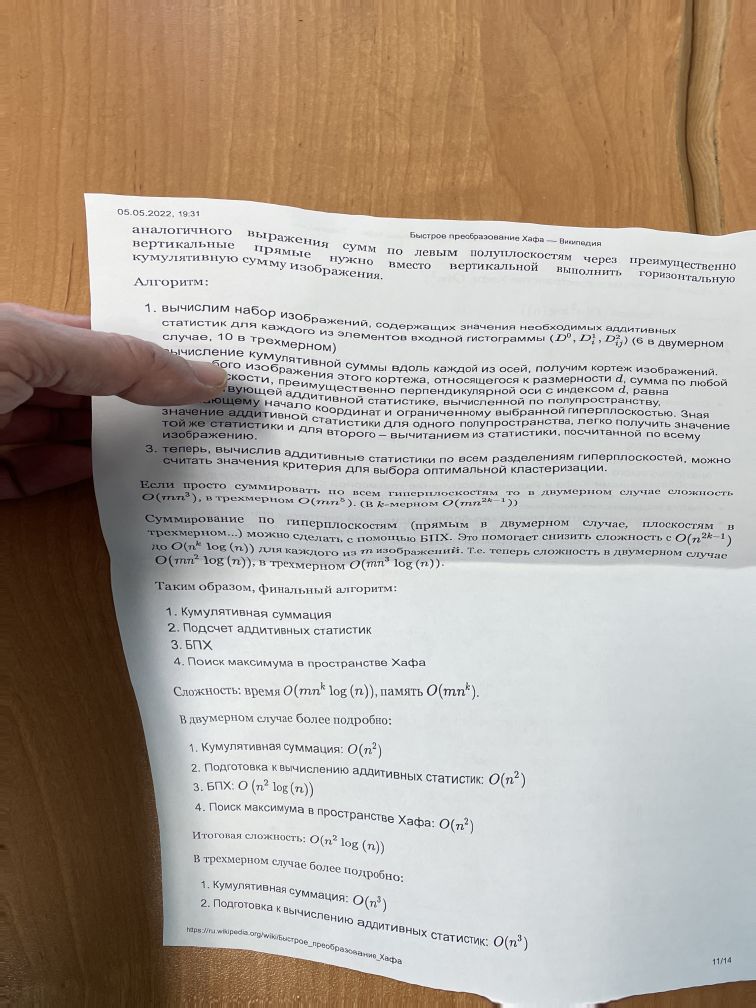}}
  \centerline{(g)}
\end{minipage}
\hfill
\begin{minipage}[b]{0.191\linewidth}
  \centering
  \centerline{\includegraphics[width=\textwidth]{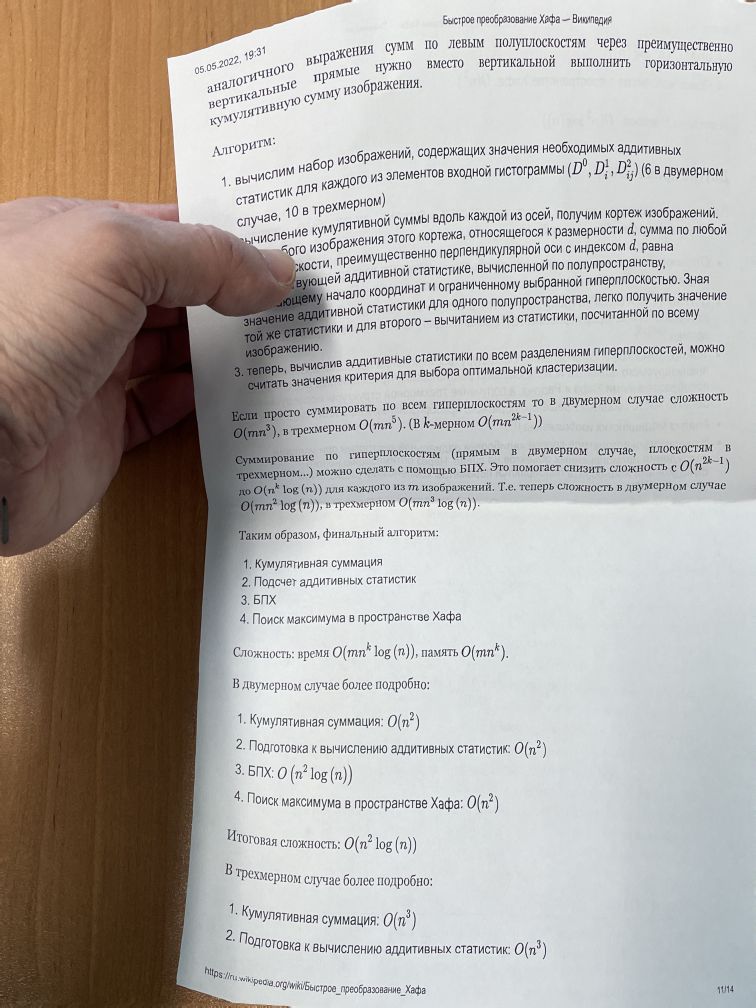}}
  \centerline{(h)}
\end{minipage}
\hfill
\begin{minipage}[b]{0.18\linewidth}
  \centering
  \centerline{\includegraphics[width=\textwidth]{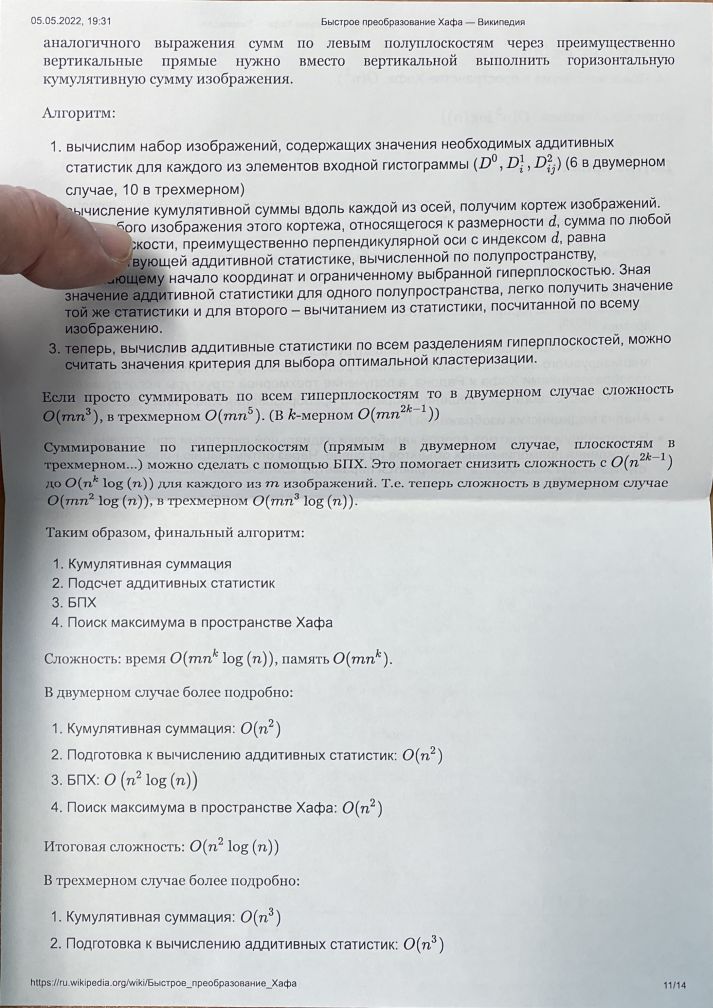}}
  \centerline{(k)}
\end{minipage}
\hfill
\begin{minipage}[b]{0.18\linewidth}
  \centering
  \centerline{\includegraphics[width=\textwidth]{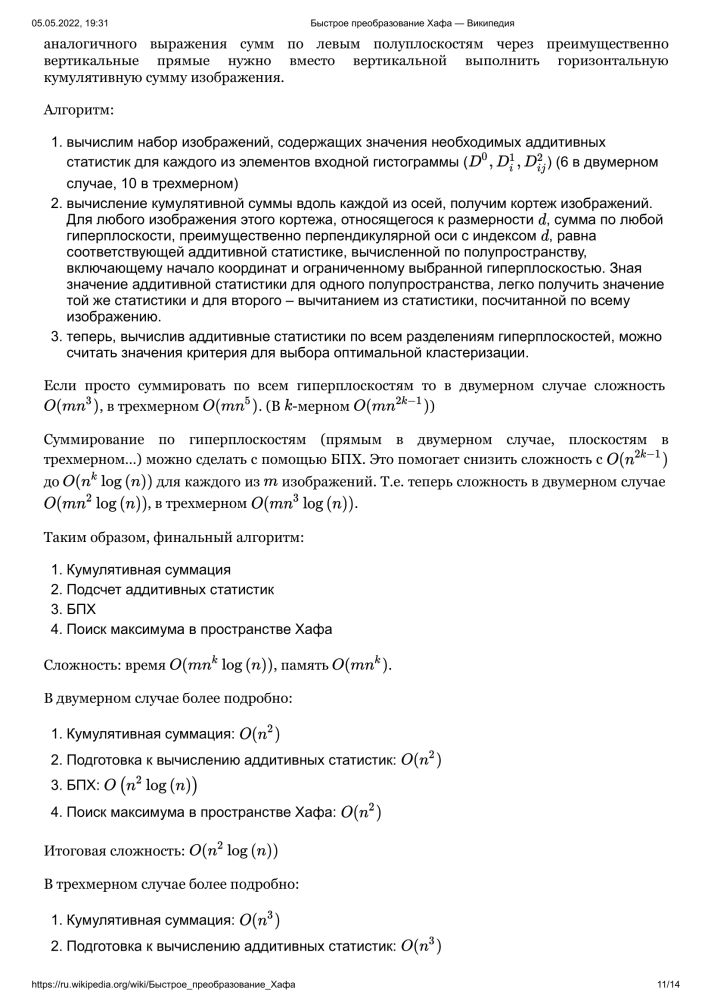}}
  \centerline{(l)}
\end{minipage}

\caption{The first row is an example from \textit{2fold/table}, the second -- \textit{2fold/hand}. The columns left to right: input, DewapNet, DocTr, Unfolder, reference.}
\label{fig:intro_img}
\end{figure}

At first let us consider rows 3 and 4. 
GeoTr shows high accuracy in the case when the document is placed on the table, though the obscuration of the document by hands has a negative impact across all metrics.
The accuracy for DewarpNet is low: the statistics of text recognition (CER and ED) are worse than even if there was no rectification algorithm applied.
Thus we can conclude that these algorithms are not robust in all cases present in \textit{2fold} subset of FDI dataset when it comes to working on \textit{raw} images.
Fig. \ref{fig:intro_img} shows that both these methods perform well on \textit{table} while having trouble rectifying images of hand-held documents.

Let us compare rows 5 and 3 for GeoTr as well as 6 and 4 for DewarpNet.
As it can be seen, cropping by a reference hexagon results in some non-trivial changes in statistics for different subsets.
The efficiency of GeoTr becomes greater on the \textit{hand} subset and poorer on the \textit{table} subset.
As for DewarpNet, it becomes more accurate in both cases.
Note that the proposed algorithm was not tested in such a setup, because cropping the image loses all the information concerning the camera parameters (focal length and principal point position), and the Unfolder relies on this information.

Based on the comparison between the row 2 and rows 3 -- 6 we can conclude that despite the fact that GeoTr is more accurate measuring by MS~SSIM and LD metrics, Unfolder shows the best accuracy in terms of character recognition.

The performance on \textit{4fold}, \textit{8fold} and \textit{3fold} is illustrated further in Table~\ref{table:multibends}.
DewarpNet and GeoTr perform uniformly across all subsets.
The Unfolder allows for the satisfactory accuracy only on \textit{4fold/table} subset.
It can be seen that the accuracy of Unfolder gradually decreases as long as document images become less and less fit for the two planes model.

An important feature of the proposed algorithm is its ability to adequately react to the document image not fit for the model of two planes, i.e. it can detect that the document image is not fit and returns trivial result.
We tested this feature on several subsets of FDI and found out that the trivial result of the Unfolder appears on 110 images of \textit{4fold} subset, 273 images of \textit{8fold} subset and 334 images of \textit{3fold} subset.
We have also tested it on the subset \textit{perspective} of the WarpDoc dataset [8] containing 170 images of non-folded rectangular documents with perspective distortions.
The Unfolder managed to get trivial result on 165 images ($97\%$).

\begin{figure}[t]
\begin{minipage}[b]{0.23\linewidth}
  \centering
  \centerline{\includegraphics[width=\textwidth]{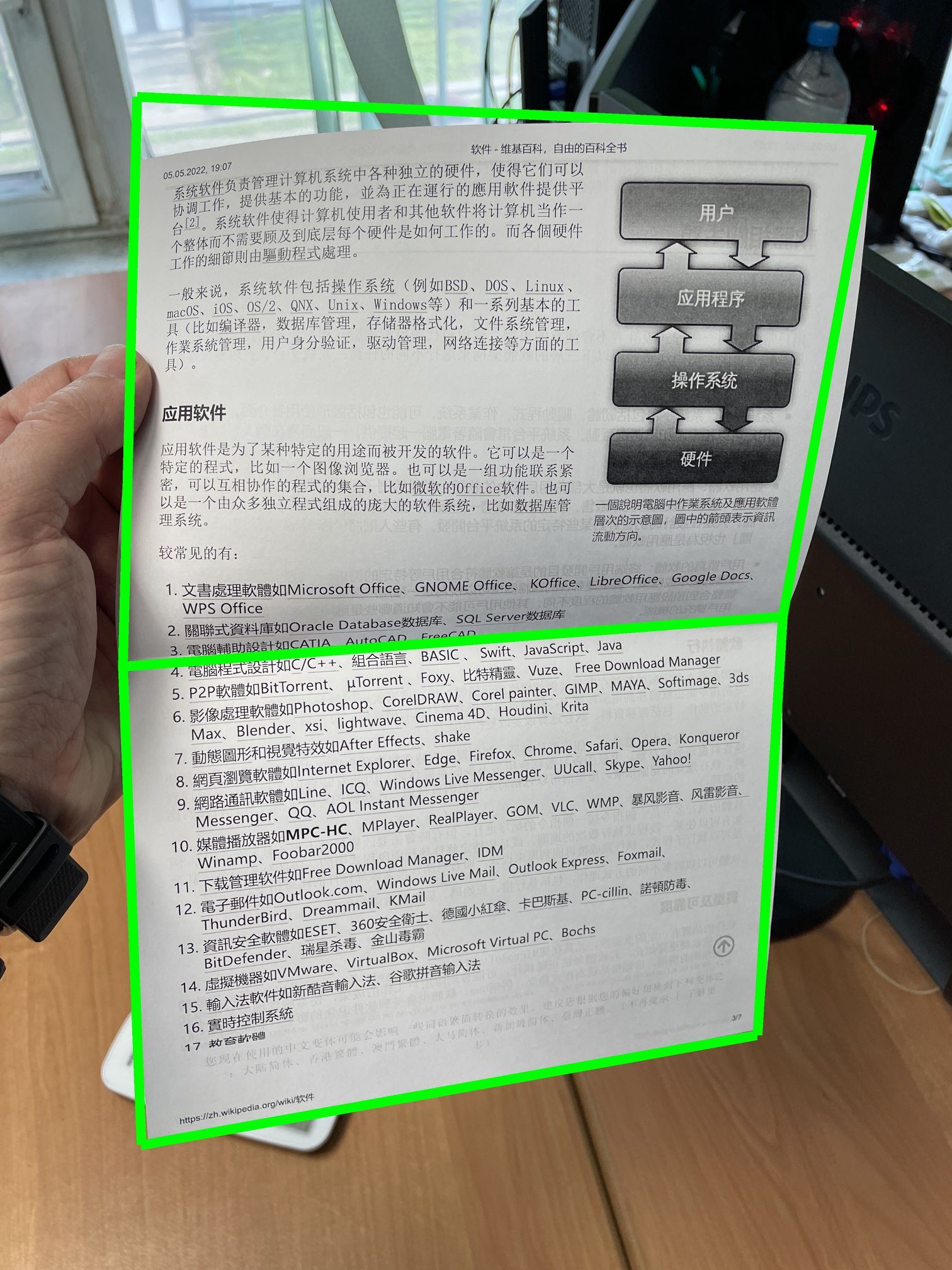}}
  \centerline{(a) }
\end{minipage}
\hfill
\begin{minipage}[b]{0.217\linewidth}
  \centering
  \centerline{\includegraphics[width=\textwidth]{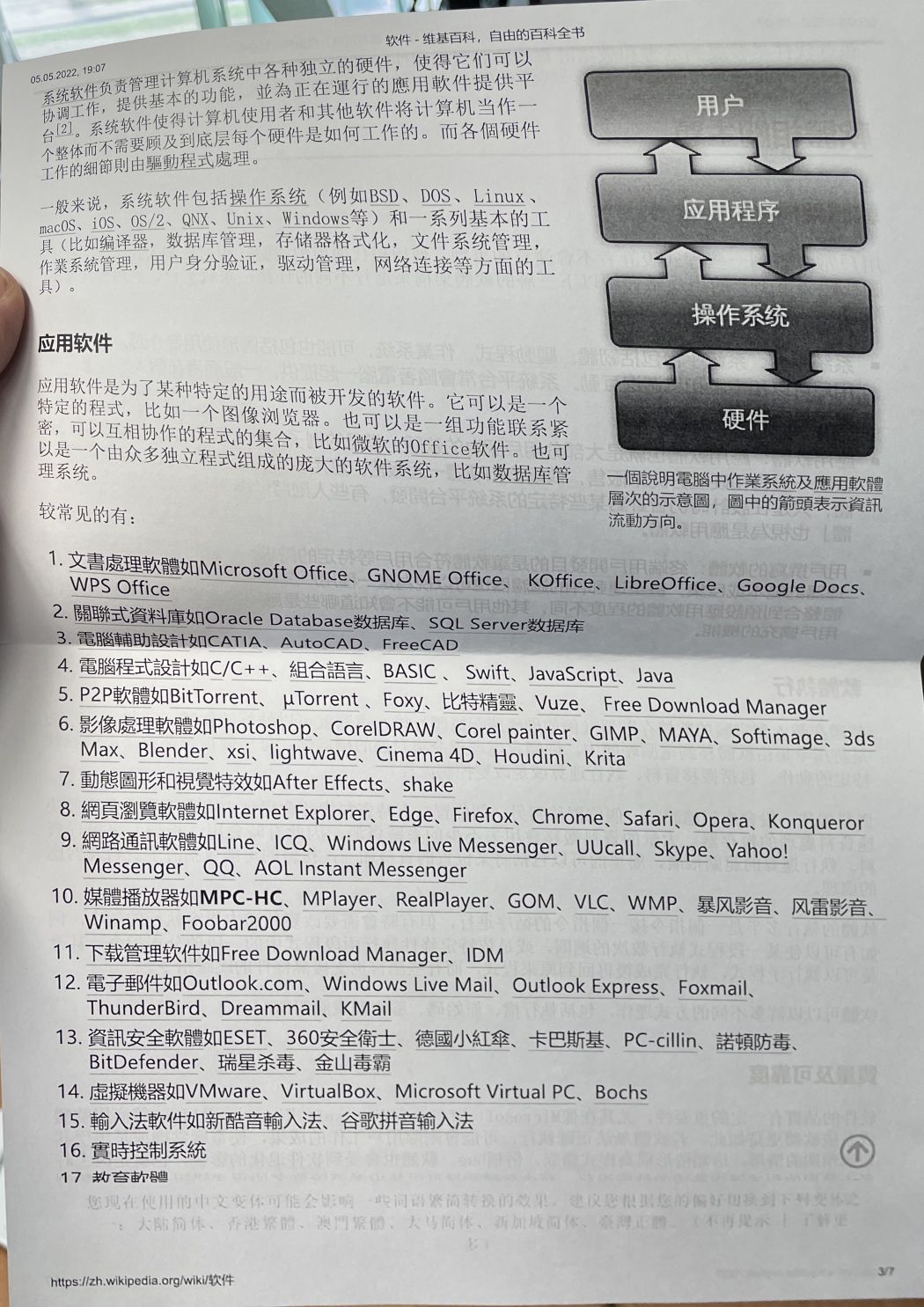}}
  \centerline{(b)}
\end{minipage}
\hfill
\begin{minipage}[b]{0.23\linewidth}
  \centering
  \centerline{\includegraphics[width=\textwidth]{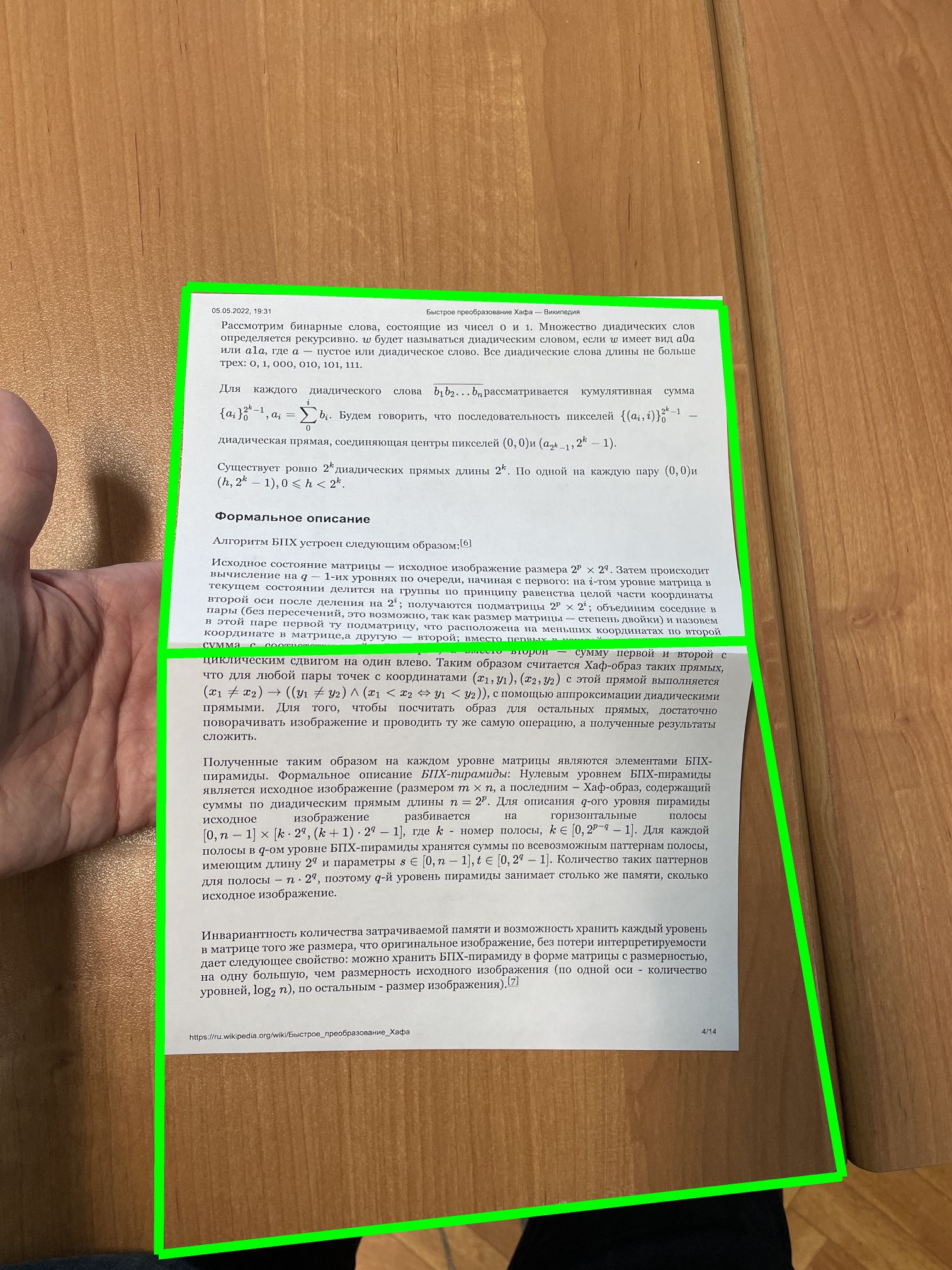}}
  \centerline{(c)}
\end{minipage}
\hfill
\begin{minipage}[b]{0.217\linewidth}
  \centering
  \centerline{\includegraphics[width=\textwidth]{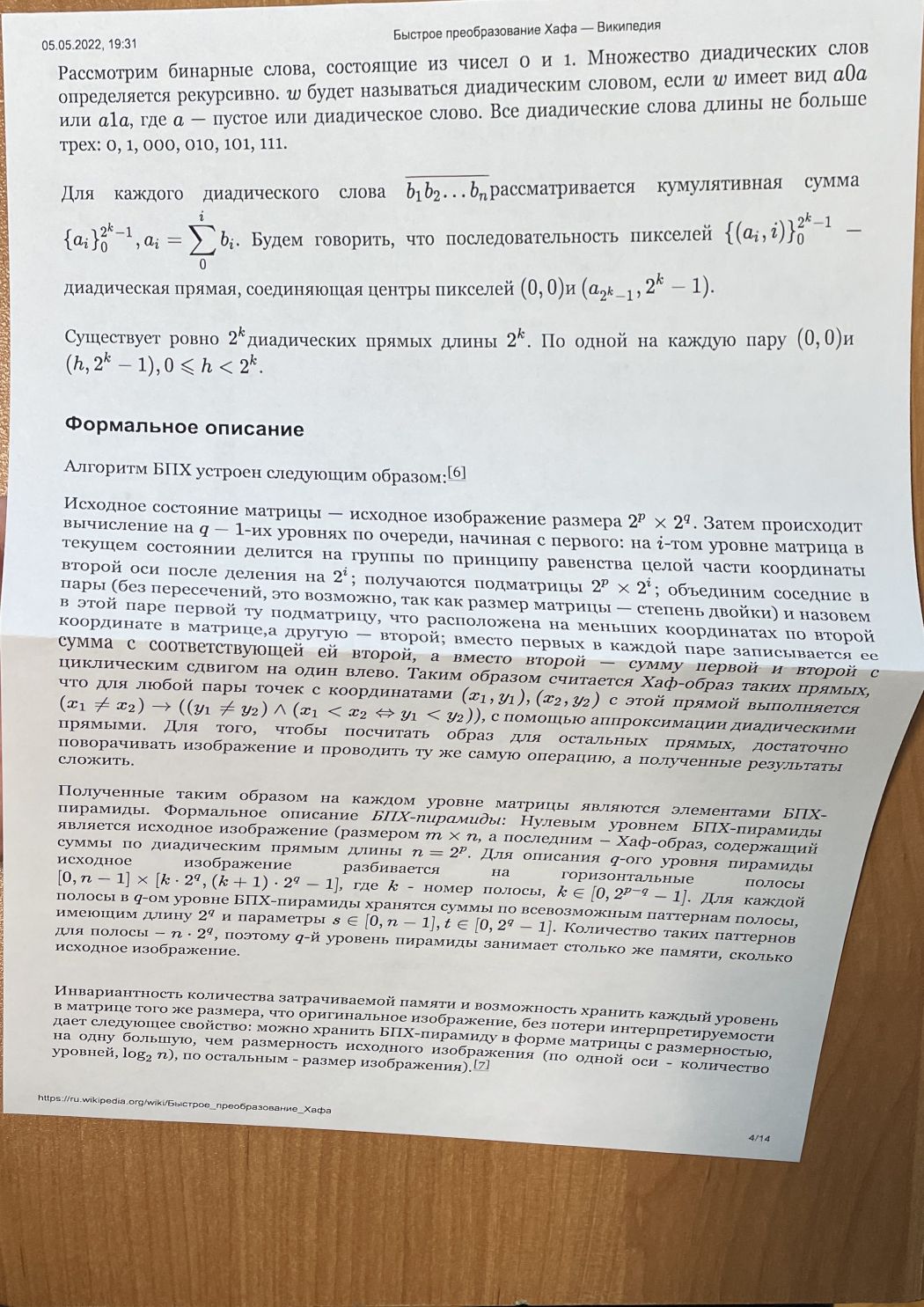}}
  \centerline{(d)}
\end{minipage}
\caption{Examples of Unfolder incorrect localization: (a, c)~input images with a detected hexangle (green), (b, d)~the corresponding rectifications
}
\label{fig:bad_cases}
\end{figure}

The Unfolder sometimes has incorrect hexangle localization resulting in bad rectification (see Fig.~\ref{fig:bad_cases}).
It also gets trivial results on 34 images of \textit{2fold} subset.
The analysis of these images shows that 20 of them ($59\%$) do not fit the two planes model (shown on Fig.~5.\textit{a}) and 13 of them ($38\%$) suffer from issues concerned with the edge detection step.

\begin{center}
	\underline{3.4 Runtime measurements}
\end{center}

Let us remind that the Unfolder algorithm includes two main steps: the localization and the projective transformation of the half pages.
The first step will be annotated with L, the second with T.

The runtime of the Unfolder was measured for three central processors in single-threaded mode: Apple A8 (iPhone 6), Apple A12 Bionic (iPhone XR), and AMD Ryzen 9 5950X (see the first row in Tab.~\ref{table:time_performance}).
The FDI subset (\textit{hand}/\textit{table} scenes in a 50/50 ratio) was employed for runtime measurements. 
In all images of this subset, the rectification of the document performed by the Unfolder was non-trivial and visually correct.  
In addition, we measured the average neural network runtime of the GeoTr (row~2) and the DewarpNet (row~3) for AMD Ryzen 9 5950X CPU and NVIDIA GeForce RTX 3090 GPU.
The resolution of the first and the last layer for DocTr pretrained model is $288 \times 288$, for DewarpNet -- $256 \times 256$.
All the CPU tests were single-threaded.

\begin{table*}[t]
\caption{The average processing time (ms) for the FDI images.  }
\centering
{
\small
\begin{tabularx}{\textwidth}{|c|Y|Y|Y|Y|}
\hline
\backslashbox{System}{Device}
  & iPhone 6 & 
   iPhone XR & 
   Desktop CPU &
   Desktop GPU 
   \\ \hline
L/T/Unfolder & 125/893/1018 & 35/216/251 & 30/84/\textbf{114} & — \\ \hline
GeoTr & — & — & 4570 & 39 \\ \hline
DewarpNet & — & — & 425 & 8 \\ \hline
\end{tabularx}
}
\label{table:time_performance}
\end{table*}

GeoTr and DewarpNet work quite fast on a desktop GPU. However, mobile GPUs are less powerful and not always available for third-party applications.
Special processing units for neural networks can be found on a small number of smartphones and also have limited availability for developers.
So, we consider the CPU as the standard device for mobile document processing.

It takes nearly 0.11 of a second to rectify an image by Unfolder on AMD Ryzen 9 5950X, a quarter of a second on the iPhone XR and nearly a second on iPhone~6.
Note, that Transformation step of Unfolder (namely projective transformation with bilinear interpolation) resulting in an image of size $2100 \times 2970$ takes from $74\%$ to $88\%$ of all processing time.
The latter can be reduced if the desired resolution of the rectified image is lowered.
However, even with this resolution Unfolder allowed for the best running time on the AMD comparing to the time of neural network execution with smaller resolution of the first and the last layer.




\begin{center}
	{\bf Conclusion}
\end{center}

We considered the problem of geometric image rectification for documents folded in half.
To account for this particular case of document distortion, we propose a two-stage algorithm Unfolder. The first stage determines the hexagonal outer edges of the document with a common horizontal vanishing point of its halves, and the second stage performs the projective transformation of each half.
We found that the common horizontal vanishing point requirement was necessary to avoid tearing the content of a document at its fold.

Along with this paper, we also publish FDI dataset which includes annotated images of documents folded in half -- twofold, as well as fourfold, eightfold, and threefold (envelope-fold) (1600 images in total). 
Two scenes were captured for each folded document: the document was placed on the table; the document was held in hand.

The Unfolder allowed for the best accuracy results on the twofold subset of the FDI dataset: the local distortion in the hand/table scenes was 15.2/7.34, the Levenshtein distance was 1158/941, and the recognition error rate was 0.35/0.3.
The proposed algorithm's average runtime was only 0.25 s per image for the iPhone XR processor and 0.11 s for the PC processor.
The comparison between state-of-the-art approach DocTr  and Unfolder demonstrates the superiority of the latter in terms of required computational resources and in terms of accuracy on the twofold subset of FDI. 
Thus, the Unfolder sets an initial baseline performance on the twofold subset of FDI.

In the future works we are planning to generalize the localization and rectification algorithms for all subsets of the FDI dataset.
Another way of Unfolder enhancement is its generalization to the case when the crease is not parallel to the short sides of the document.

\begin{center}
	{\bf References}
\end{center}

[1] Arlazarov VV, Zhukovsky A, Krivtsov V, Nikolaev D, Polevoy D. Analysis of Using Stationary and Mobile Small-Scale Digital Video Cameras for Document Recognition [In Russian]. Information Technologies and Computation Systems 2014; (3): 71-78.

[2] Burie J, Chazalon J, Coustaty M, Eskenazi S, Luqman MM, Mehri M, Nayef N, Ogier J, Prum S, Rusi{\~n}ol M. ICDAR2015 competition on smartphone document capture and OCR (SmartDoc). In: 2015 13th International Conference on Document Analysis and Recognition (ICDAR): 1161-1165. IEEE(2015). DOI: 10.1109/ICDAR.2015.7333943.

[3] Hartl A, Reitmayr G. Rectangular target extraction for mobile augmented reality applications. In: Proceedings of the 21st International Conference on Pattern Recognition (ICPR2012): 81-84. IEEE(2012).

[4] Puybareau E, G{\'e}raud T. Real-time document detection in smartphone videos. In: 2018 25th IEEE International Conference on Image Processing (ICIP): 1498-1502. IEEE(2018). DOI: 10.1109/ICIP.2018.8451533.

[5] Tropin DV, Ershov AM, Nikolaev DP, Arlazarov VV. Advanced Hough-based method for on-device document localization. Computer Optics 2021; 45 (5): 702--712. DOI: 10.18287/2412-6179-CO-895.

[6] Das S, Mishra G, Sudharshana A, Shilkrot R. The common fold: utilizing the four-fold to dewarp printed documents from a single image. In: Proceedings of the 2017 ACM Symposium on Document Engineering: 125--128. (2017). DOI: 10.1145/3103010.3121030.

[7] Ma K, Shu Z, Bai X, Wang J, Samaras D. DocUNet: Document image unwarping via a stacked u-net. In: Proceedings of the IEEE conference on computer vision and pattern recognition: 4700--4709. (2018). DOI: 10.1109/CVPR.2018.00494.

[8] Xue C, Tian Z, Zhan F, Lu S, Bai S. Fourier Document Restoration for Robust Document Dewarping and Recognition. In: Proceedings of the IEEE/CVF Conference on Computer Vision and Pattern Recognition: 4573--4582. (2022). DOI: 10.1109/CVPR52688.2022.00453.

[9] Tan CL, Zhang L, Zhang Z, Xia T. Restoring warped document images through 3d shape modeling. IEEE Transactions on Pattern Analysis and Machine Intelligence 2005; 28 (2): 195--208. DOI: 10.1109/TPAMI.2006.40.

[10] Zhang L, Yip AM, Brown MS, Tan CL. A unified framework for document restoration using inpainting and shape-from-shading. Pattern Recognition 2009; 42 (11): 2961--2978. DOI: 10.1016/j.patcog.2009.03.025.

[11] You S, Matsushita Y, Sinha S, Bou Y, Ikeuchi K. Multiview Rectification of Folded Documents. IEEE Transactions on Pattern Analysis and Machine Intelligence 2018; 40 (2): 505-511. DOI: 10.1109/TPAMI.2017.2675980.

[12] Luo D, Bo P. Geometric Rectification of Creased Document Images based on Isometric Mapping. arXiv preprint arXiv:2212.08365 2022.

[13] Brown MS, Seales WB. Image restoration of arbitrarily warped documents. IEEE Transactions on pattern analysis and machine intelligence 2004; 26 (10): 1295--1306. DOI: 10.1109/TPAMI.2004.87.

[14] Zhang L, Zhang Y, Tan C. An Improved Physically-Based Method for Geometric Restoration of Distorted Document Images. IEEE Transactions on Pattern Analysis and Machine Intelligence 2008; 30 (4): 728-734. DOI: 10.1109/TPAMI.2007.70831.

[15] Sun M, Yang R, Yun L, Landon G, Seales B, Brown MS. Geometric and photometric restoration of distorted documents. In: Tenth IEEE International Conference on Computer Vision (ICCV'05) Volume 1: 1117--1123. IEEE(2005). DOI: 10.1109/ICCV.2005.106.

[16] Meng G, Wang Y, Qu S, Xiang S, Pan C. Active flattening of curved document images via two structured beams. In: Proceedings of the IEEE Conference on Computer Vision and Pattern Recognition: 3890--3897. (2014). DOI: 10.1109/CVPR.2014.497.

[17] Brown MS, Tsoi Y. Geometric and shading correction for images of printed materials using boundary. IEEE Transactions on Image Processing 2006; 15 (6): 1544--1554. DOI: 10.1109/tip.2006.871082.

[18] Koo HI, Cho NI. Rectification of figures and photos in document images using bounding box interface. In: 2010 IEEE Computer Society Conference on Computer Vision and Pattern Recognition: 3121--3128. IEEE(2010). DOI: 10.1109/CVPR.2010.5540071.

[19] Tsoi Y, Brown MS. Multi-view document rectification using boundary. In: 2007 IEEE Conference on Computer Vision and Pattern Recognition: 1--8. IEEE(2007). DOI: 10.1109/CVPR.2007.383251.

[20] Coons SA. Surfaces for computer-aided design of space forms. 1967. Technical report, MIT/LCS/TR-41.

[21] Stamatopoulos N,  Gatos B, Pratikakis I, Perantonis SJ. A two-step dewarping of camera document images. In: 2008 The Eighth IAPR International Workshop on Document Analysis Systems: 09--216. IEEE(2008). DOI: 10.1109/DAS.2008.40.

[22] Gaofeng M, Chunhong P, Shiming X, Jiangyong D, Nanning Z. Metric Rectification of Curved Document Images. IEEE Transactions on Pattern Analysis and Machine Intelligence 2012; 34 (4): 707--722. DOI: 10.1109/TPAMI.2011.151.

[23] Fu B, Wu M, Li R, Li W, Xu Z, Yang C. A model-based book dewarping method using text line detection. In: Proc. 2nd Int. Workshop on Camera Based Document Analysis and Recognition, Curitiba, Barazil: 63--70 (2007).

[24] Das S, Ma K, Shu Z, Samaras D, Shilkrot R. DewarpNet: Single-Image Document Unwarping With Stacked 3D and 2D Regression Networks. In: Proceedings of International Conference on Computer Vision: 131--140. (2019). DOI: 10.1109/ICCV.2019.00022.

[25] Feng H, Zhou W, Deng J, Tian Q, Li H. DocScanner: Robust Document Image Rectification with Progressive Learning. arXiv preprint arXiv:2110.14968; 2021.

[26] Feng H, Wang Y, Zhou W, Deng J, Li H. DocTr: Document Image Transformer for Geometric Unwarping and Illumination Correction. In: Proceedings of the 29th ACM International Conference on Multimedia: 273--281. (2021). DOI: 10.48550/arXiv.2110.12942.

[27] Das S, Singh KY, Wu J, Bas E, Mahadevan V, Bhotika R, Samaras D. End-to-end Piece-wise Unwarping of Document Images. In: Proceedings of the IEEE/CVF International Conference on Computer Vision: 4268--4277. (2021). DOI: 10.1109/ICCV48922.2021.00423.

[28] Xie G, Yin F, Zhang X, Liu C. Document Dewarping with Control Points. In: International Conference on Document Analysis and Recognition: 466--480. Springer(2021). DOI: 10.1007/978-3-030-86549-8\_30.

[29] Jiang X, Long R, Xue N, Yang Z, Yao C, Xia G. Revisiting Document Image Dewarping by Grid Regularization. In: Proceedings of the IEEE/CVF Conference on Computer Vision and Pattern Recognition: 4543--4552. (2022). DOI: 10.1109/CVPR52688.2022.00450.

[30] Wang Y, Zhou W, Lu Z, Li H. UDoc-GAN: Unpaired Document Illumination Correction with Background Light Prior. In: Proceedings of the 30th ACM International Conference on Multimedia: 5074--5082. (2022). DOI: 10.1145/3503161.3547916.

[31] Ma K, Das S, Shu Z, Samaras D. Learning From Documents in the Wild to Improve Document Unwarping. In: ACM SIGGRAPH 2022 Conference Proceedings: 1--9 (2022). DOI: 10.1145/3528233.3530756.

[32] Feng H, Zhou W, Deng J, Wang Y, Li H. Geometric Representation Learning for Document Image Rectification. In: European Conference on Computer Vision: 475--492. Springer(2022). DOI: 10.1007/978-3-031-19836-6\_27.

[33] Das S, Ma K, Shu Z, Samaras D. Learning an Isometric Surface Parameterization for Texture Unwrapping. In: European Conference on Computer Vision: 580--597. Springer(2022). DOI: 10.1007/978-3-031-19836-6\_33.

[34] Li X, Zhang B, Liao J, Sander PV. Document rectification and illumination correction using a patch-based CNN. ACM Transactions on Graphics (TOG) 2019; 38 (6): 1--11. DOI: 10.1145/3355089.3356563.

[35] Bandyopadhyay H, Dasgupta T, Das N, Nasipuri M. A gated and bifurcated stacked u-net module for document image dewarping. In: 2020 25th International Conference on Pattern Recognition (ICPR): 10548--10554. IEEE(2021). DOI: 10.1109/ICPR48806.2021.9413001.

[36] Xie G, Yin F, Zhang X, Liu C. Dewarping document image by displacement flow estimation with fully convolutional network. In: International Workshop on Document Analysis Systems: 131--144. Springer(2020). DOI: 10.1007/978-3-030-57058-3\_10.

[37] Xu Z, Yin F, Yang P, Liu C. Document Image Rectification in Complex Scene Using Stacked Siamese Networks. In: 2022 26th International Conference on Pattern Recognition (ICPR): 1550--1556. IEEE(2022). DOI: 10.1109/ICPR56361.2022.9956331.

[38] Verhoeven F, Magne T, Sorkine-Hornung O. Neural Document Unwarping using Coupled Grids. arXiv preprint arXiv:2302.02887 2023. DOI: 10.48550/arXiv.2302.02887.

[39] Feng H, Liu S, Deng J, Zhou W, Li H. Deep Unrestricted Document Image Rectification. arXiv preprint arXiv:2304.08796 2023. DOI: 10.48550/arXiv.2304.08796.

[40] Hertlein F, Naumann A, Philipp P. Inv3D: a high-resolution 3D invoice dataset for template-guided single-image document unwarping. International Journal on Document Analysis and Recognition (IJDAR) 2023: 1--12. DOI: 10.1007/s10032-023-00434-x.

[41] Brady ML. A fast discrete approximation algorithm for the Radon transform. SIAM Journal on Computing 1998; 27 (1): 107--119. DOI: 10.1137/S0097539793256673.

[42] Shemiakina J, Konovalenko I, Tropin D, Faradjev I. Fast projective image rectification for planar objects with Manhattan structure. In: Twelfth International Conference on Machine Vision (ICMV 2019): 450--458. SPIE(2020). DOI: 10.1117/12.2559630.

[43] Skoryukina N, Nikolaev DP, Sheshkus A, Polevoy D. Real time rectangular document detection on mobile devices. In: Seventh International Conference on Machine Vision (ICMV 2014): 458--463. SPIE(2015). DOI: 10.1117/12.2181377.

[44] Zhang Z, He L. Whiteboard scanning and image enhancement. Digital signal processing 2007; 17 (2): 414--432. DOI: 10.1016/j.dsp.2006.05.006.

[45] Trusov A, Limonova E. The analysis of projective transformation algorithms for image recognition on mobile devices. In: Twelfth International Conference on Machine Vision (ICMV 2019): 250--257. SPIE(2020). DOI: 10.1117/12.2559732.

[46] Dutta A, Zisserman A. The VIA Annotation Software for Images, Audio and Video. In: Proceedings of the 27th ACM International Conference on Multimedia: 2276–2279. ACM(2019). DOI: 10.1145/3343031.3350535.

[47] Wang Z, Simoncelli EP, Bovik AC. Multiscale structural similarity for image quality assessment. In: The Thrity-Seventh Asilomar Conference on Signals, Systems \& Computers, 2003: 1398--1402. IEEE(2003). DOI: 10.1109/ACSSC.2003.1292216.

[48] Liu C, Yuen J, Torralba A, Sivic J, Freeman WT. Sift flow: Dense correspondence across different scenes. In: European conference on computer vision: 28--42. Springer(2008). DOI: 10.1007/978-3-540-88690-7\_3.

[49] Levenshtein VI and others. Binary codes capable of correcting deletions, insertions, and reversals. In: Soviet physics doklady: 707--710 (1966).

[50] Smith R. An overview of the Tesseract OCR engine. In: Ninth international conference on document analysis and recognition (ICDAR 2007): 629--633. IEEE(2007). DOI: 10.1109/ICDAR.2007.4376991.

\begin{center}
	{\bf Supplementary materials. Appendix A.}
\end{center} 

{\bf Proposition.} 
Let there be two quadrilaterals on the plane: $A_1B_1CD$ and $A_2B_2CD$ with the adjacent side $CD$, located in different half-planes relative to $CD$. 
Let there be rectangles $A_1'B_1'C'D'$ and $A_2'B_2'C'D'$ with the adjacent side $C'D'$ in different half-planes relative to it. 
Let $H_1$ be a projective transformation from $A_1B_1CD$ to $A_1'B_1'C'D'$, and $H_2$ be a projective transformation from $A_2B_2CD$ to $A_2'B_2'C'D'$.
Let $H$ be a mapping equal to $H_1$ on the half-plane with $A_1B_1CD$ and equal to $H_2$ on the half-plane with $A_2B_2CD$.
Then $H$ is correctly defined   on the segment $CD$ if and only if the intersecting lines $A_1B_1, A_2B_2$ and $CD$ share a common point.

{\bf Proof.} 
Consider an arbitrary point $X$ on $CD$.
Denote $X'_1 = H_1(X),\, X'_2 = H_2(X)$.
Denote $V_1 = A_1B_1 \cap CD,\, V_2 = A_2B_2 \cap CD$.
Since the projective transform preserves the cross ratio,
\[
\frac{CD}{XD} : \frac{CV_1}{DV_1} = \frac{C'D'}{X'_1D'} : \frac{C'V_1'}{D'V_1'} \iff
X'_1D' = \frac{C'D' \cdot C'V_1'}{D'V_1'} \cdot \frac{CV_1}{DV_1} : \frac{CD}{XD}
\]
\[
\frac{CD}{XD} : \frac{CV_2}{DV_2} = \frac{C'D'}{X'_2D'} : \frac{C'V_2'}{D'V_2'} \iff
X'_2D' = \frac{C'D' \cdot C'V_2'}{D'V_2'} \cdot \frac{CV_2}{DV_2} : \frac{CD}{XD}
\]

Since $A_1'B_1'C'D'$ and $A_2'B_2'C'D'$ are rectangles, $V_1' = V_2' = \infty$, so the equality $X'_1D = X'_2D$ is equivalent to the equality $V_1 = V_2$.

\begin{center}
	{\bf Author’s information}
\end{center} 

{\bf Alexandr Mikhailovich Ershov}, (b. 1999), Ph.D student in the Institute for Information Transmission Problems, graduated from Lomonosov Moscow State University in 2023, is a developer at Smart Engines Service LLC since 2020. Research interests: image processing and computer vision.\\
E-mail: {\it a.ershov@smartengines.com}

{\bf Daniil Vyacheslavovich Tropin}, (b. 1995), Ph.D. in Computer Science, is a developer in Smart Engines Service LLC, a junior researcher in Federal Research Center “Computer Science and Control”, Russian Academy of Sciences. 
Research interests: image processing and computer vision. \\
E-mail: {\it daniil\_tropin@smartengines.com}

{\bf Elena Evgenievna Limonova}, obtained Master's degree from Moscow Institute of Physics and Technology in 2017. Since 2016, Elena has worked as a programmer and technician at Smart Engines Service LLC, and since 2022 as a researcher at Federal Research Center “Computer Science and Control” of Russian Academy of Sciences. In 2023 she obtained Ph.D. degree in Computer Science. Research interests: neural network compression and efficient image recognition on mobile devices. \\
E-mail: {\it limonova@smartengines.com}

{\bf Dmitry Petrovich Nikolaev}, (b. 1978), graduated from Lomonosov Moscow State University in 2000, Ph.D. in Physics and Mathematics, is a head of the vision systems laboratory at the Institute for Information Transmission Problems, CTE of Smart Engines Service LLC. Research interests: machine vision, algorithms for fast image processing, pattern recognition. \\
E-mail: {\it dimonstr@iitp.ru}

{\bf Vladimir Viktorovich Arlazarov},  (b. 1976) received a specialist degree in Applied Mathematics from the Moscow Institute of Steel and Alloys in 1999, Ph.D. in Computer Science, is a head of division 93 at the Federal Research Center ``Computer Science and Control'',  Russian Academy of Sciences, CEO of Smart Engines Service LLC.
Research interests: computer vision and document analysis systems. \\
E-mail: {\it vva@smartengines.com}

\begin{center}
\textit{Code of State Categories Scientific and Technical Information (in Russian – GRNTI)): 28.23.15}

\textit{Received XXX. The final version -- YYY.}
\end{center}

\end{document}